\documentclass{article}

\PassOptionsToPackage{numbers, compress, sort}{natbib}
\usepackage[preprint]{neurips_2025}  

\usepackage[utf8]{inputenc} 
\usepackage[T1]{fontenc}    
\usepackage{hyperref}       
\usepackage{url}            
\usepackage{booktabs}       
\usepackage{amsfonts}       
\usepackage{nicefrac}       
\usepackage{microtype}      
\usepackage{xcolor}         

\usepackage{float}
\usepackage{stfloats}
\usepackage{graphicx}
\usepackage{algorithm}
\usepackage[noend]{algpseudocode}
\usepackage{amsthm,amsmath, amssymb}
\usepackage{booktabs, ctable}
\usepackage{multirow}

\title{One-Time Soft Alignment Enables \\ Resilient Learning without Weight Transport}

\author{
  Jeonghwan Cheon$^{1}$ \quad Jaehyuk Bae$^{1}$ \quad Se-Bum Paik$^{1}$\\
  $^1$Department of Brain and Cognitive Sciences \\
  Korea Advanced Institute of Science and Technology, Daejeon, Republic of Korea \\
  \texttt{\{jeonghwan518, jaehyukbae, sbpaik\}@kaist.ac.kr}\\
}

\begin{document}

\maketitle
\vspace{-0.3cm}
\begin{abstract}
  Backpropagation is the cornerstone of deep learning, but its reliance on symmetric weight transport and global synchronization makes it computationally expensive and biologically implausible. Feedback alignment offers a promising alternative by approximating error gradients through fixed random feedback, thereby avoiding symmetric weight transport. However, this approach often struggles with poor learning performance and instability, especially in deep networks. Here, we show that a one-time soft alignment between forward and feedback weights at initialization enables deep networks to achieve performance comparable to backpropagation, without requiring weight transport during learning. This simple initialization condition guides stable error minimization in the loss landscape, improving network trainability. Spectral analyses further reveal that initial alignment promotes smoother gradient flow and convergence to flatter minima, resulting in better generalization and robustness. Notably, we also find that allowing moderate deviations from exact weight symmetry can improve adversarial robustness compared to standard backpropagation. These findings demonstrate that a simple initialization strategy can enable effective learning in deep networks in a biologically plausible and resource-efficient manner.
\end{abstract}
\vspace{-0.3cm}

\section{Introduction}

Recent breakthroughs in artificial intelligence have been driven by a powerful learning rule, backpropagation \cite{rumelhart1986, lecun2015}. Leveraging this method, modern large-scale deep networks have achieved superhuman performance in domains such as image recognition \cite{krizhevsky2012, he2016} and language modeling \cite{achiam2023, team2023}. Despite its effectiveness, backpropagation faces significant challenges in terms of computational and resource efficiency \cite{strubell2020, sze2017, schwartz2020}. For instance, training a single large language model can consume over a thousand megawatt‐hours of electricity \cite{patterson2021} and emit tens of tons of carbon dioxide \cite{luccioni2023}, raising concerns about its sustainability. This heavy energy consumption largely stems from a core limitation of backpropagation: it requires dynamic access to separate memory units during learning  \cite{strubell2020, horowitz2014, chen2016}.

In contrast to backpropagation on von Neumann architectures \cite{turing1936, von1945} —which requires dynamic memory access between separate compute and storage units — the biological brain learns through locally connected, physically fixed neurons \cite{markram2015}. This fundamental difference raises two key questions: How does the brain learn efficiently using only local connections? And how can we replicate such energy-efficient learning in artificial systems? These questions have motivated interest in biologically plausible deep learning and neuromorphic computing as promising alternatives \cite{indiveri2015, merolla2014}. Unlike backpropagation, which depends on symmetric forward and backward pathways, biological neural circuits rely on distinct feedforward and feedback connections \cite{felleman1991, bastos2015}. However, how the brain learns using such asymmetric pathways remains an open challenge, known as the weight transport problem \cite{grossberg1987, crick1989, lillicrap2016, lillicrap2020, richards2019}.

Notably, feedback alignment algorithms address the weight transport problem by using fixed, random synaptic feedback \cite{lillicrap2016}. It demonstrates that networks can learn without explicitly copying forward weights into the backward pass. During training, forward weights gradually “align” with the fixed random feedback weights, approximating the error signals of backpropagation. While this offers a biologically plausible mechanism with separate feedforward and feedback pathways, its performance — especially in deep networks — remains significantly below that of standard backpropagation and fails to explain the brain’s remarkable learning efficiency \cite{bartunov2018}.

Recent approaches have proposed biologically plausible learning rules that begin to close this performance gap, yet they still grapple with the weight transport problem. For example, the sign-symmetry algorithm communicates only the sign of each weight, rather than its full value, but still requires sign information to be transmitted during learning \cite{liao2016, xiao2018}. Predictive coding \cite{whittington2017, song2024} and equilibrium propagation \cite{scellier2017} provide compelling energy-based frameworks but rely on perfectly symmetric forward and backward weights — effectively assuming exact synaptic mirroring. Other methods, such as weight mirror and the Kolen–Pollack algorithm \cite{kolen1994, akrout2019} adapt the feedback pathway to promote weight alignment, but at the cost of additional bidirectional connections — an arrangement that lacks biological plausibility. In short, these methods partially replace weight transport with constrained or adaptive feedback, but still depend on sign transmission, tied weights, or added connectivity throughout learning.

Here, we propose a different approach to biologically plausible learning, focusing on soft alignment at initialization rather than enforcing symmetry throughout training. We show that aligning forward and backward weights once at initialization, without any weight transport thereafter, is sufficient to guide learning. This initial alignment shapes the learning dynamics, enabling convergence to a smoother loss landscape and improving trainability, generalization, and robustness. Furthermore, our method demonstrates superior adversarial robustness compared to standard backpropagation. 

Overall, these results identify a minimal yet effective requirement for biologically plausible networks to approach backpropagation-level performance, while offering a computationally efficient solution to the credit assignment problem in neural computation.

\section{Preliminaries}

\subsection{Credit assignment and backpropagation}

To optimize a network to perform a given task — to enable learning — each neuron must adjust its synaptic weights in a way that reduces the discrepancy between its actual and expected outputs. This, in turn, requires assigning an appropriate portion of the overall error to each weight, a challenge known as the credit assignment problem. In machine learning, backpropagation \cite{rumelhart1986} offers an effective solution to this problem and has become the standard method for training artificial neural networks. 

Concretely, consider a typical multi-layer feedforward network for pattern classification, denoted as $f_\theta: \mathbb{R}^m \rightarrow \mathbb{R}^d$, parameterized by $\theta = \{\mathbf{W}_l, \mathbf{b}_l\}_{l=0}^{L-1}$. Given an input $\mathbf{x} \in \mathbb{R}^m$, the network computes its output through the forward pass:

\vspace{-0.35cm}
\begin{equation}
\label{eq:forward}
    \mathbf{o}_{l+1} = \mathbf{W}_l \mathbf{h}_l + \mathbf{b}_l, \quad \mathbf{h}_{l+1} = \phi(\mathbf{o}_{l+1}),
\end{equation}
\vspace{-0.35cm}

where $\mathbf{W}_l$ denotes the forward weight matrix at layer $l$, $\mathbf{b}_l$ is the corresponding bias vector. The signal is propagated through a nonlinear activation function $\phi$. Once the forward pass is complete, the loss is computed as the discrepancy between the network’s output and the ground-truth labels. Backpropagation then calculates the gradient of this loss with respect to each weight and bias, and these gradients are subsequently used to update the network parameters.

\vspace{-0.35cm}
\begin{equation}
\label{eq:backprop}
\quad \delta_l = \frac{\partial \mathcal{L}}{\partial \mathbf{o}_l} = (\mathbf{W}_l^T \delta_{l+1}) \odot \phi'(\mathbf{o}_l),
\end{equation}
\vspace{-0.35cm}

where $\delta_l$ denotes the error signal at layer $l$, $\phi'$ represents the derivative of the activation function, and $\odot$ indicates the element-wise product. Based on this error signal, the weights are updated as

\vspace{-0.35cm}
\begin{equation}
\label{eq:weight-update}
\Delta \mathbf{W}_l = -\eta \delta_{l+1} \mathbf{h}_l^T,
\end{equation}
\vspace{-0.35cm}

where $\eta$ denotes the learning rate. 

\subsection{Weight transport problem and feedback alignment}

Although backpropagation endows neural networks with powerful learning capabilities, it relies on exact knowledge of the forward weights in the next layer to compute the backward error signal — known as the weight transport problem \cite{grossberg1987,crick1989} — which underlies both its energy inefficiency and its lack of biological plausibility. 

Feedback alignment \cite{lillicrap2016} offers a more biologically plausible alternative by replacing the backward‐pathway weights with fixed, random matrices $\mathbf{B}_l$, thereby eliminating the need for weight transport:

\vspace{-0.35cm}
\begin{equation}
    \label{eq:fa}
    \delta_l = \frac{\partial \mathcal{L}}{\partial \mathbf{o}_l} = (\mathbf{B}_l \delta_{l+1}) \odot \phi'(\mathbf{o}_l).
\end{equation}
\vspace{-0.35cm}

Since feedback alignment substitutes the core backpropagation operation — the use of the transpose of the forward weights $\mathbf{W}_l$ — with a random, fixed matrix $\mathbf{B}_l$, the resulting error signal is initially uninformative. However, during training, the forward weights $\mathbf{W}_l$ gradually adapt to “align” with the fixed $\mathbf{B}_l^\top$. This emergent alignment causes the error signals (\ref{eq:fa}) to approximate those computed by backpropagation (\ref{eq:backprop}).

\section{One-time initial feedback alignment}

\algnewcommand\algorithmicforeach{\textbf{for each}}
\algdef{S}[FOR]{ForEach}[1]{\algorithmicforeach\ #1\ \algorithmicdo}

\begin{algorithm}[ht]
\caption{Feedback alignment with initial weight alignment}\label{alg1}
\begin{algorithmic}[1]
\Procedure {Initialization with soft initial alignment}{$network$ $f_\theta, \theta_{init} $}
    \For {layer $l$ = $L$-1 \textbf{to} $0$}
        \State $\mathbf{B}_{l} \sim \mathcal{N}(0, \frac{a^2}{fan_{in}})$ \Comment{initialize backward weights}
        \State $\mathbf{W}_{l} = \mathbf{B}_{l}^\top\cos(\theta_{init}) + \mathbf{R} \sin(\theta_{init}), \mathbf{R} \sim \mathcal{N}(0, \frac{a^2}{fan_{in}})$
        \Comment{initialize forward weights}
    \EndFor
\EndProcedure

\Procedure {feedback alignment}{$network$ $f_\theta, \text{data}$}
    \ForEach {epoch}
        \ForEach {$\text{batch} = (\mathbf{x}, \mathbf{y}) \in \text{data}$}
        \State $\mathcal{L}(\theta) = \text{Loss}(f_\theta(\mathbf{x}), \mathbf{y})$ \Comment{forward pass, equation (\ref{eq:forward})}
        \State $\delta_L = \frac{\partial \mathcal{L}}{\partial \mathbf{o}_L}$ \Comment{compute error}
        \For {layer $l$ = $L$-1 \textbf{to} $0$}
        \State $\mathbf{W}_{l} = \mathbf{W}_{l} -\eta\,\boldsymbol{\delta}_{l+1}\,\mathbf{h}_{l}^T$ \Comment{update weights, equation (\ref{eq:weight-update})}
        \State $\boldsymbol{\delta}_{l}= (\mathbf{B}_{l}\, \boldsymbol{\delta}_{l+1}) \odot \phi'(\mathbf{o}_l)\ $ \Comment{compute error, equation (\ref{eq:fa})}
        \EndFor
    \EndFor
    \EndFor
\EndProcedure
\end{algorithmic}
\end{algorithm}

Even though feedback alignment eliminates the need for exact weight transport, its performance remains significantly inferior to that of standard backpropagation. In this work, we propose that imposing a soft initial alignment between forward and backward weights can dramatically improve biologically plausible deep learning. Unlike earlier schemes that rely on sign-constrained feedback or maintain a strict coupling between forward and backward weights throughout training, our method requires soft alignment only at initialization, after which the forward and backward weights are allowed to evolve independently. Concretely, we initialize the forward weights $\mathbf{W}_l$ to match the fixed random backward weights $\mathbf{B}_l$ directly or align with them in a “soft” sense via a projection (Algorithm \ref{alg1}). In contrast to baseline feedback alignment, our method begins training with an aligned configuration between forward and backward weights—yet it never requires further synchronization during learning.

We proposed initial alignment as a means to improve neural computation without requiring explicit weight transport during learning. Importantly, there are biologically plausible mechanisms through which such an initialization condition could arise during development. Recent studies have shown that mimicking the brain’s developmental process \cite{galli1988, ackman2012, anton2019, martini2021}, for example, through pretraining with random noise can yield weight alignment, thereby providing a “soft” alignment when the network begins learning from real data \cite{cheon2024}. Similarly, other work has demonstrated that sign symmetry can emerge through ligand‐specific axonal guidance during wiring of forward‐ and feedback‐projecting neurons \cite{liao2016}. Although initial alignment itself may not be strictly biologically plausible, these findings suggest that such alignment could arise naturally during development.

\vspace{-0.1cm}
\section{Results}
\vspace{-0.1cm}

\vspace{-0.1cm}
\subsection{One-time initial alignment enables learning without weight transport}
\vspace{-0.1cm}

\begin{figure}[t!]
	\centering
	\includegraphics[width=\textwidth]{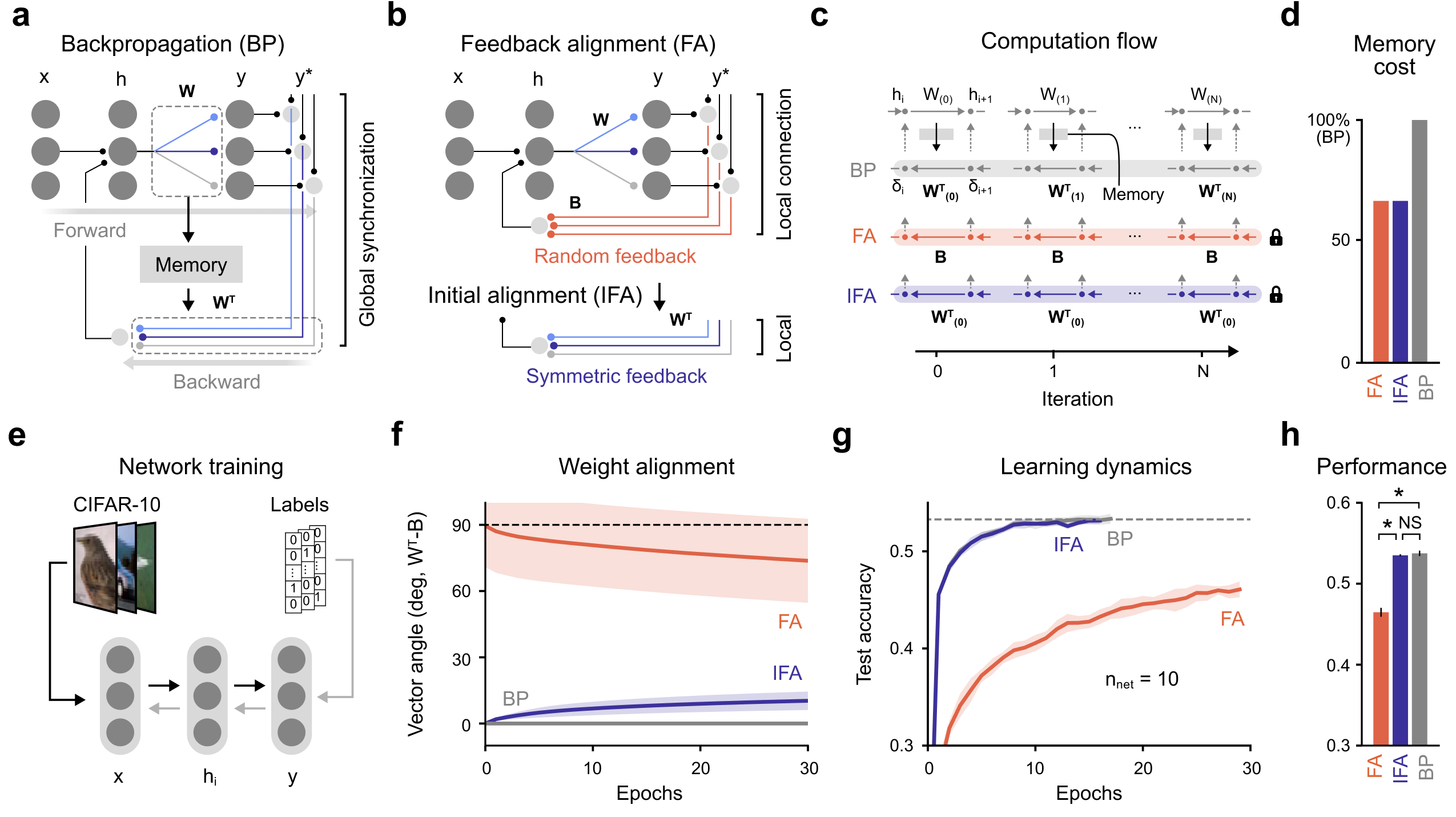}
    \vspace{-0.5cm}
    \caption{Effect of initial weight alignment on learning without weight transport.
        (a) Schematic of the backpropagation (BP) algorithm, where forward and backward weights are kept synchronized during training through external memory access.
        (b) Schematic of feedback alignment (FA) algorithms, which use a separate backward pathway with fixed, random feedback weights. Initial feedback alignment (IFA) also uses fixed feedback weights but aligns them with the forward weights only at initialization.
        (c) Computation flow for implementing backpropagation and feedback alignment on‐chip.
        (d) Comparison of memory access cost between FA and BP.
        (e) Neural network models trained on the CIFAR-10 dataset using various learning algorithms.
        (f) Alignment angle between forward and backward weights in the final layer of the network.
        (g) Learning curves showing test accuracy over the course of training.
        (h) Final test accuracy.
    }
	\label{fig1}
\end{figure}

To compare the effect of initial alignment, we investigated neural network learning under three different feedback dynamics while keeping all other conditions constant. Backpropagation (BP) is the standard training algorithm, in which the forward weights are copied to the backward path at every iteration to compute gradients (Figure \ref{fig1}a). We also tested feedback alignment (FA), a biologically plausible alternative that avoids the weight transport problem by employing fixed, random feedback weights (Figure \ref{fig1}b). Our proposed model, initial feedback alignment (IFA), adopts the same feedback pathway as FA but differs at initialization: the backward weights are set equal to the forward weights at the start of training (Figure \ref{fig1}c). Like FA, IFA requires no weight transport during training and can be implemented with local circuits, and it requires a one‐time initial alignment. We compared the computational flow of these three feedback dynamics (Figure \ref{fig1}d). In BP, forward weights must be copied into memory and accessed during each backward pass. In contrast, both FA and IFA use fixed feedback weights throughout training, eliminating the need for per‐iteration weight transport or memory access. As a result, their memory‐access overhead is significantly lower than that of BP (Figure \ref{fig1}d).

To compare these learning algorithms, we trained multi-layer feedforward networks on a natural image classification task using three distinct feedback schemes (Figure \ref{fig1}e). To characterize their learning dynamics, we first measured the alignment between forward and backward weights (Figure \ref{fig1}f) by computing the cosine similarity of corresponding weight vectors. As expected, BP maintains exact symmetry: forward and backward weights are identical because each forward weight is stored and retrieved during the backward pass. In FA, fixed random feedback weights gradually induce alignment in the forward weights during training. In contrast, IFA begins with forward weights set equal to the backward weights and then gradually relaxes this alignment during learning. We next examined the learning curves for each feedback scheme (Figure \ref{fig1}g). Remarkably, IFA’s convergence trajectory closely mirrors that of BP: it learns substantially faster than baseline FA and ultimately achieves performance comparable to BP. Indeed, final accuracy under IFA is significantly higher than under FA (Figure \ref{fig1}g, right; FA vs. IFA, $n_{\text{net}}=10$, two‐sided rank‐sum test, $P<10^{-3}$) and not significantly different from BP (BP vs. IFA, $n_{\text{net}}=10$, two‐sided rank‐sum test, NS, $P=0.082$). These findings suggest that a one-time alignment of forward and feedback weights at initialization is sufficient to enable learning performance comparable to BP, without requiring ongoing synchronization throughout training.

In addition, we evaluated test accuracy under various conditions to assess the effect of initial weight alignment. Using several standard image classification benchmarks — including SVHN \cite{netzer2011}, CIFAR-10, CIFAR-100 \cite{krizhevsky2009} and STL-10 \cite{coates2011} — we found that initializing forward and backward weights consistently leads to improved test accuracy. We further extended our experiments to convolutional neural network (CNN) architectures \cite{lecun1998, krizhevsky2012}, confirming that this initialization strategy enhances learning performance even without any weight‐transport during training.

\vspace{-0.1cm}
\subsection{Stabilizing learning through initial weight alignment}
\vspace{-0.1cm}

\begin{figure}[h!]
	\centering
	\includegraphics[width=\textwidth]{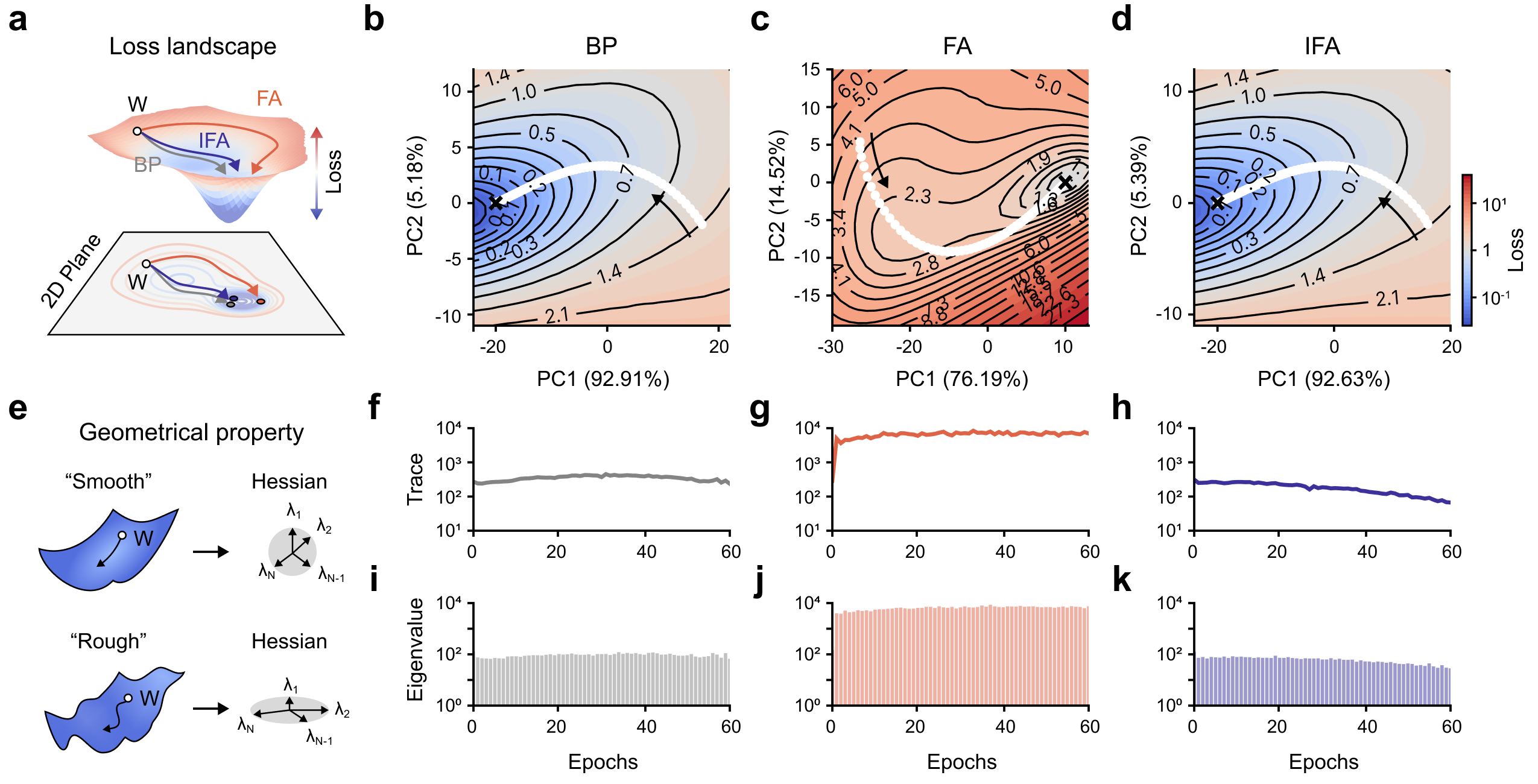}
    \vspace{-0.5cm}
	\caption{Loss landscape analysis of learning trajectories under different training rules.
        (a) Illustration of neural networks traversing the loss landscape during training, with each trajectory corresponding to a different learning algorithm.
        (b-d) Training trajectories projected onto a two-dimensional subspace defined by the first two principal components (PCA): (b) BP, (c) FA, (d) IFA.
        (e) Geometric properties of the loss landscape, quantified using the Hessian’s spectrum; smoother landscapes exhibit narrower spectra (i.e., smaller spectral radius).
        (f-h) Trace of the Hessian over the course of training.
        (i-k) Maximum eigenvalue of the Hessian during training for BP, FA, and IFA, respectively.
    }
	\label{fig2}
\end{figure}

Next, we examined how initial alignment influences learning dynamics within the loss landscape, using principal component analysis \cite{li2018} (Figure \ref{fig2}a) to visualize training trajectories. Specifically, we compared the trajectories produced under three feedback dynamics: BP (Figure \ref{fig2}b), FA (Figure \ref{fig2}c), and IFA (Figure \ref{fig2}d). BP exhibits stable and directed trajectories that effectively converge toward the global minimum. In contrast, FA results in more erratic trajectories — longer and less direct — indicating inefficient navigation of the loss landscape. Notably, IFA produces stable trajectories and a loss landscape structure that closely resemble those of BP, suggesting that initial alignment facilitates more efficient and stable learning.

To quantitatively compare the learning trajectories under different feedback dynamics, we performed a spectral analysis of the loss Hessian \cite{yao2020}. Specifically, we assessed the smoothness of the loss landscape (Figure \ref{fig2}e) by measuring the trace of the Hessian (Figure \ref{fig2}f–h) and its maximum eigenvalue (Figure \ref{fig2}i–k). Consistent with our trajectory-based observations, BP maintains low Hessian trace and top eigenvalue throughout training, indicating a smooth loss surface. In contrast, FA exhibits substantially higher trace and top eigenvalue, suggesting a much rougher landscape. Notably, IFA preserves low Hessian trace and top eigenvalues across the entire learning process. These results suggest that providing initial guidance for error minimization is sufficient to promote stable learning dynamics and a smooth optimization trajectory, even in the absence of ongoing weight transport.

\begin{figure}[t!]
	\centering
	\includegraphics[width=\textwidth]{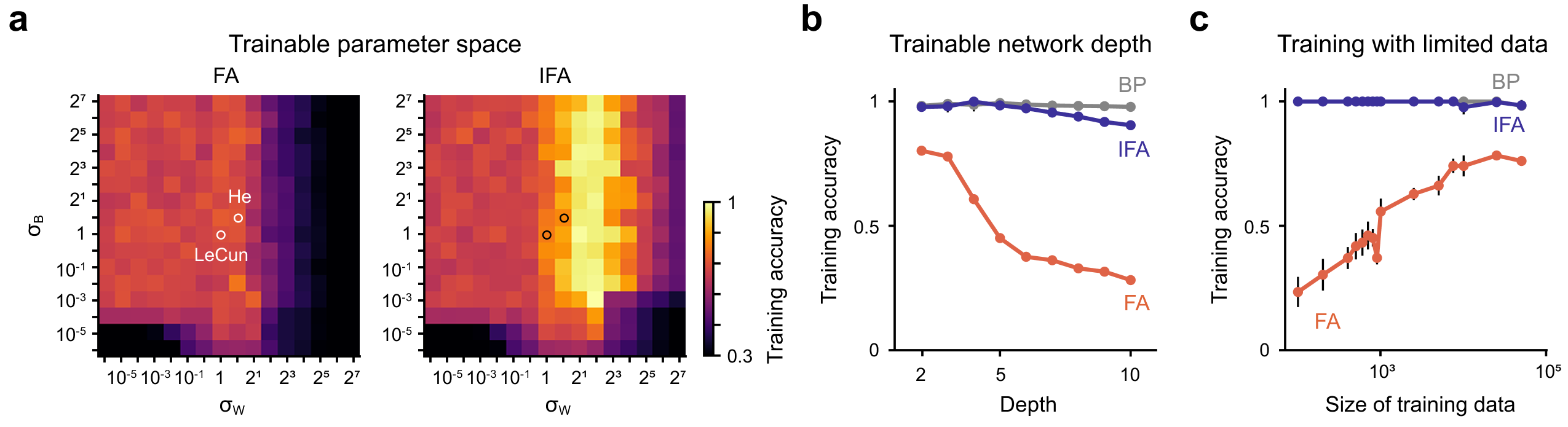}
    \vspace{-0.5cm}
    \caption{Trainability of neural networks under various conditions.
        (a) Trainable parameter space: the x- and y-axes represent the variances of the forward and backward weights, respectively. Color indicates the final training accuracy. The point $(1, 1)$ corresponds to LeCun initialization; $(\sqrt{2}, \sqrt{2})$ corresponds to He initialization.  
        (b) Trainability versus network depth: final training accuracy for feedforward networks with depths ranging from 2 to 10 layers.  
        (c) Trainability under limited data: final training accuracy as a function of training set sizes, ranging from 100 to 50,000 samples.
    }
	\label{fig3}
\end{figure}

Building on the observation that stable learning induced by initial alignment between forward and backward weights governs the entire training process, we hypothesized that this alignment enhances neural network trainability under various conditions. First, we investigated networks initialized with different combinations of forward and backward weight variances (Figure \ref{fig3}a). 3 Next, we examined how initial alignment influences the maximum trainable depth (Figure \ref{fig3}b). In baseline FA, trainability rapidly degrades as depth increases, whereas IFA enables training at greater depths, comparable to those achievable with backpropagation. Finally, we considered a limited-data regime (Figure \ref{fig3}c). Because FA must learn to approximate backpropagation from data, it fails when the dataset is too small. In contrast, IFA enables successful training even under severe data constraints, achieving performance comparable to BP.

\vspace{-0.1cm}
\subsection{Initial alignment enhances robust generalization}
\vspace{-0.1cm}

\begin{figure}[h!]
	\centering
	\includegraphics[width=\textwidth]{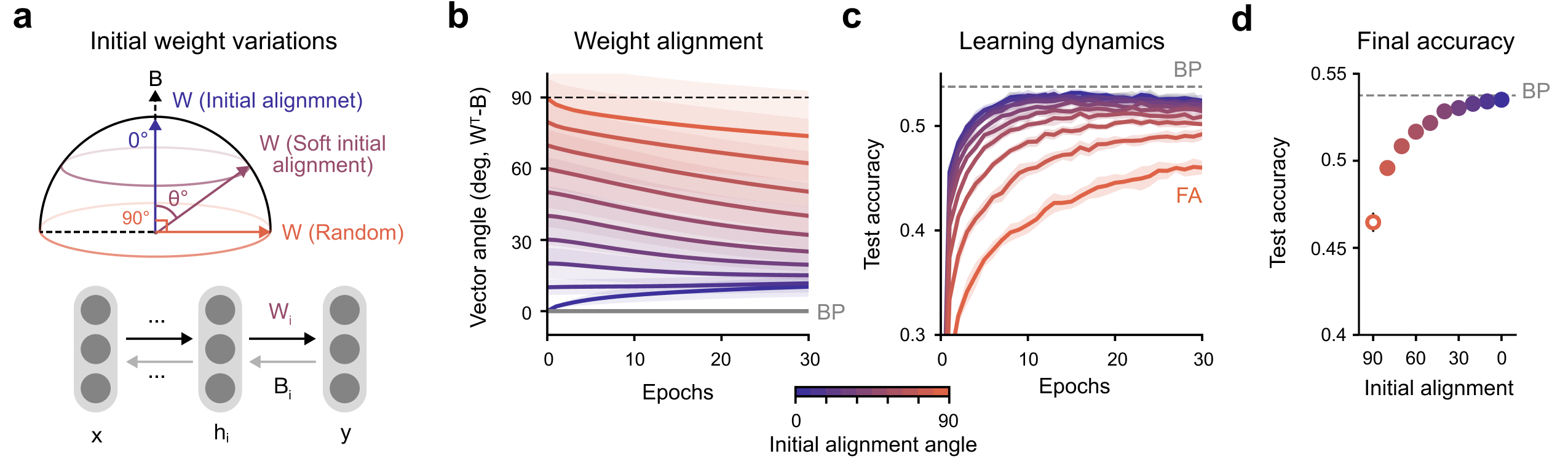}
    \vspace{-0.5cm}
	\caption{Effect of the degree of initial alignment on network performance.
        (a) Schematic of initialization conditions for forward and backward weights in feedback alignment. Perfect initial alignment sets the backward weights equal to the forward weights at initialization, whereas baseline feedback alignment uses randomly initialized backward weights. Soft alignment is achieved by sampling forward weights within a subspace defined by a specified angle relative to the backward weights.
        (b) Angle between forward and backward weights over the course of training.
        (c) Learning curves showing test accuracy during training.
        (d) Relationship between the initial alignment angle and final test accuracy.
        }
	\label{fig4}
\end{figure}

To evaluate the impact of initial alignment, we systematically varied the alignment angle between forward and backward weights and analyzed the resulting learning dynamics (Figure \ref{fig4}a). An initial angle of 0° indicates perfect alignment, where forward and backward weights are identical, while an angle of 90° corresponds to orthogonal weights, as in baseline feedback alignment. Intermediate (“soft”) alignments were generated by projecting the forward weights onto a subspace defined by a specific angle relative to the backward weights. We found that the learning dynamics depend strongly on the initial alignment (Figure \ref{fig4}b). When the initial alignment is strong, learning proceeds by gradually relaxing this alignment; when the initial alignment is weak, the network learns by progressively aligning the weights. The learning curves reflect this behavior (Figure \ref{fig4}c): stronger initial alignment leads to faster convergence and higher final test accuracy. Plotting the final test accuracy as a function of the initial alignment angle (Figure \ref{fig4}d) reveals a trend — performance improves with stronger alignment but eventually saturates. These results suggest that while a good initial alignment is critical for effective learning, perfect symmetry between forward and backward weights is not necessary.

\begin{figure}[h!]
	\centering
	\includegraphics[width=\textwidth]{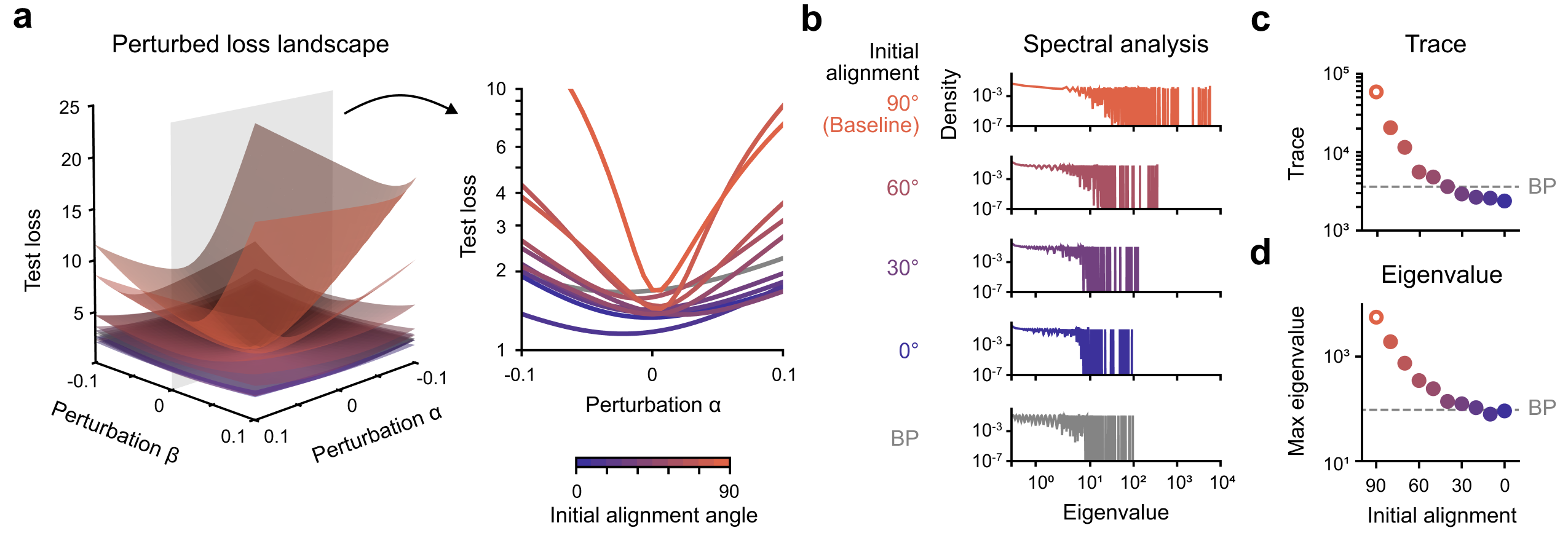}
    \vspace{-0.5cm}
	\caption{Spectral analysis of convergence across initial alignment angles.
        (a) Loss landscape visualized by perturbing the trained model parameters within a two-dimensional subspace defined by the top two eigenvectors of the Hessian. The axes, denoted by $\alpha$ and $\beta$, represent scaling factors for perturbations along the first and second eigenvector directions, respectively. The landscapes are color-coded according to each network’s initial alignment angle.
        (b) Spectral density of the Hessian eigenvalues, illustrating the distribution of curvature across the parameter space.
        (c) Relationship between the initial alignment angle and the trace of the Hessian.
        (d) Relationship between the initial alignment angle and the largest Hessian eigenvalue.
        }
	\label{fig5}
\end{figure}

Next, we performed a spectral analysis of the converged optima obtained under varying initial alignment angles. Specifically, we visualized the loss landscape by perturbing the model parameters along two principal directions (Figure \ref{fig5}a), defined by the eigenvectors corresponding to the largest and second-largest eigenvalues of the Hessian. The baseline FA method exhibits a rough loss landscape, where even small perturbations in parameter space lead to significant increases in loss. In contrast, applying a soft initial alignment dramatically smooths the loss landscape of the converged optima, making it comparable to that observed with backpropagation. Furthermore, increasing the degree of initial weight alignment leads to progressively smoother loss landscapes.

Additionally, we validated this trend by analyzing the spectral density of Hessian eigenvalues (Figure \ref{fig5}b). Under baseline FA, the spectrum is shifted toward higher eigenvalues and exhibits greater dispersion compared to BP. Introducing initial alignment compresses the spectrum, resulting in a distribution that closely matches that of BP. To quantify the smoothness of the loss landscape, we computed both the Hessian trace (Figure \ref{fig5}c) and its largest eigenvalue (Figure \ref{fig5}d). Both metrics decrease as the initial alignment angle increases, and then plateau beyond a certain alignment threshold. These geometric insights align with our earlier observation that test accuracy improves and then saturates with stronger initial alignment (Figure \ref{fig4}d). Collectively, these findings indicate that initial weight alignment plays a critical role in shaping a smooth loss landscape — a known factor in promoting robust generalization \cite{hochreiter1997, keskar2016, chaudhari2019, izmailov2018, foret2021}. Notably, only a moderate degree of alignment is necessary to obtain these benefits.

\begin{figure}[h!]
	\centering
	\includegraphics[width=\textwidth]{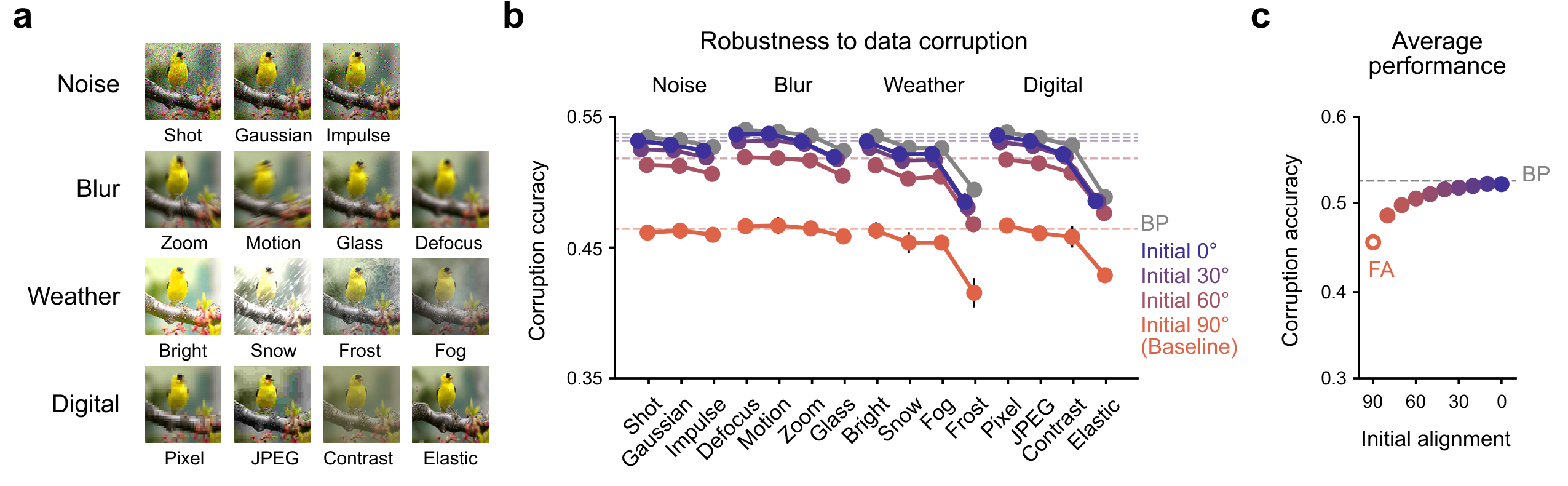}
    \vspace{-0.5cm}
	\caption{Effect of initial weight alignment on corruption robustness.
        (a) Fifteen image corruption types, grouped into four categories — noise, blur, weather, and digital — were applied to test samples to evaluate network robustness.
        (b) Classification accuracy on corrupted inputs (solid lines with markers) compared to accuracy on clean, uncorrupted images (dotted lines).
        (c) Relationship between the initial alignment angle and robustness, measured as the average accuracy across all corruption types.
    }
	\label{fig6}
\end{figure}

The spectral analysis revealed that initial weight alignment promotes convergence to a smoother loss landscape. Prior work has suggested that smoother loss landscape reduce a network's sensitivity to input distortions or perturbations \cite{foret2021}, implying that initial weight alignment may enhance robustness. To test this hypothesis, we emplyoed a benchmark image dataset featuring fifteen corruption types, categorized into noise, blur, weather, and digital distortions \cite{hendrycks2018} (Figure \ref{fig6}a). We trained networks using feedback dynamics across varying initial alignment angles and evaluated their classification accuracy on the corrupted inputs (Figure \ref{fig6}b). Notably, increasing the degree of initial alignment consistently improved test accuracy across all corruption types (Figure \ref{fig6}c). Furthermore, when the severity of each corruption was gradually increased, networks with stronger initial alignment exhibited greater resistance to these progressively challenging distortions. These results suggest that initial weight alignment, by promoting a smoother loss landscape, ultimately enhances the robustness of the network.

\vspace{-0.1cm}
\subsection{The role of weight misalignment in promoting adversarial robustness}
\vspace{-0.1cm}

\begin{figure}[h!]
	\centering
	\includegraphics[width=\textwidth]{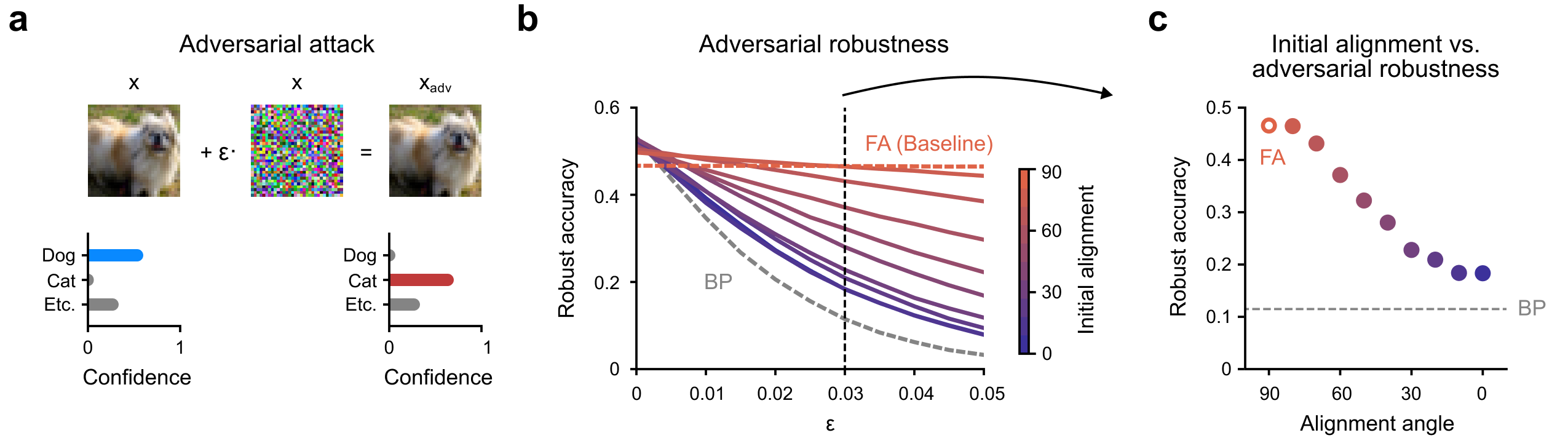}
    \vspace{-0.5cm}
	\caption{Effect of weight misalignment on adversarial robustness.
        (a) Illustration of an adversarial attack: small, targeted perturbations are added to input images to induce misclassification.
        (b) Classification accuracy as a function of increasing adversarial perturbation strength.
        (c) Relationship between the initial alignment angle and adversarial robustness, measured as classification accuracy under attack with perturbation magnitude $\epsilon = 0.03$.
        }
	\label{fig7}
\end{figure}

Thus far, we have emphasized that stronger initial weight alignment enhances standard training performance. Conversely, we also demonstrate that weight misalignment can confer benefits in the context of adversarial robustness. To evaluate this, we conducted adversarial attacks using the Fast Gradient Sign Method (FGSM) (Figure \ref{fig7}a), which applies gradient‐based perturbations to input images to induce misclassification \cite{goodfellow2014}. Under standard BP, model accuracy rapidly deteriorates as the perturbation strength increases (Figure \ref{fig7}b, BP). In contrast, models trained with IFA maintain significantly higher accuracy, even under stronger attacks (Figure \ref{fig7}b, IFA). This robustness appears to emerge from the gradual “loosening” of weight alignment that naturally occurs during fixed‐feedback learning, which reduces the impact of adversarial perturbations. Furthermore, we observed that larger initial misalignment further improves adversarial robustness (Figure \ref{fig7}c). Taken together, these findings highlight a functional advantage of weight misalignment and underscore the practical benefits of employing “soft” alignment at initialization.

\section{Discussion}
\vspace{-0.1cm}

We demonstrated that initial soft alignment can effectively guide backpropagation-like learning, even without enforcing weight transport between forward and backward pathways throughout training. Specifically, soft alignment at initialization facilitates stable convergence to flatter optima, leading to improved generalization and robustness. Interestingly, we also identified a functional benefit of misalignment: enhanced adversarial robustness compared to standard backpropagation. Overall, these findings present a simple yet powerful pathway toward biologically plausible deep learning.

\textbf{Critical role of initial conditions in learning dynamics.} Traditional backpropagation algorithms rely on exact symmetry between forward and backward weights throughout training. In contrast, we show that soft alignment imposed only at initialization is sufficient to achieve backpropagation‐like learning in terms of performance, robustness, and stability. This finding suggests that a one-time, approximate alignment of weights at the start of training can support error-driven learning without requiring explicit weight transport during the training process. Supporting this view, prior studies have shown that initialization plays a pivotal role in neural network behavior, influencing accuracy \cite{glorot2010, sutskever2013, he2015, saxe2013} and calibration \cite{cheon2024b}. Collectively, these results highlight the fundamental role of initial conditions in shaping the learning dynamics and functional development of neural networks.

\textbf{Developmental foundations of biologically plausible weight alignment.} Unlike artificial neural networks, which typically require externally provided datasets for training, the biological brain begins learning even before birth as it is exposed to sensory inputs \cite{galli1988, ackman2012, anton2019, martini2021}. During this prenatal stage, spontaneous neural activity — often resembling random noise — drives the formation and refinement of synaptic connections. Notably, prior work has shown that pretraining neural networks using random noise inputs can promote alignment between forward and backward weights under feedback alignment algorithm \cite{cheon2024}. This suggests that spontaneous activity–dependent refinement may serve as a developmental mechanism for establishing soft initial weight alignment. Furthermore, cell‐type–specific axonal guidance mediated by chemical ligands may contribute to forming feedforward and feedback circuits with initial symmetry in weight signs \cite{liao2016}.

\textbf{Initialization strategies for neuromorphic computing.} In neuromorphic hardware systems \cite{indiveri2015, merolla2014} and physical neural networks \cite{wright2022, momeni2023}, initializing weight alignment between forward and backward pathways offers a simple yet effective strategy to support learning. From an implementation perspective, this approach imposes minimal overhead: backward weights can be initialized as a mirror of forward weights directly within on-chip memory, requiring no additional memory allocation or computation overhead. Despite its simplicity, our findings suggest that this initialization method can promote stable learning dynamics and achieve performance comparable to standard backpropagation, without incurring costly weight transport or frequent memory read/write operations typically required.

\vspace{-0.1cm}
\section{Broader impacts and limitations}
\vspace{-0.1cm}

\textbf{Broader impacts.} One of the most pressing challenges in artificial intelligence today is energy inefficiency. As model sizes increase and AI systems are deployed across a wide range of applications, the associated electricity consumption and carbon emissions have raised serious concerns about the long-term sustainability of current approaches. Biologically plausible learning rules that avoid explicit weight transport offer a promising path toward more energy‐efficient computation. Our findings can be leveraged in neuromorphic hardware to reduce power consumption and environmental impact, thereby contributing to the development of more sustainable AI technologies.

\textbf{Limitations.} This study primarily focused on multi‐layer feedforward networks and evaluated performance using standard benchmark tasks. We also extended our experiments to convolutional architectures and found that initial weight alignment consistently improves the performance of feedback alignment. However, a performance gap remains between our approach and full backpropagation in deeper or more complex models. Therefore, while our method offers a simple and hardware‐friendly mechanism to improve learning without requiring weight transport, it does not yet match the accuracy of conventional backpropagation.

\vspace{-0.1cm}
\section{Code availability}
\vspace{-0.1cm}
Python 3.12 (Python Software Foundation) with PyTorch 2.1 was used to perform the simulation. The code used in this work will be made available after the paper is published.

\newpage
\begin{ack}
This work was supported by the National Research Foundation of Korea (NRF) grants (NRF2022R1A2C3008991 to S.P.) and by the Singularity Professor Research Project of KAIST (to S.P.).
\end{ack}

\medskip
{
\small
\bibliographystyle{unsrt}
\bibliography{references}
}

\newpage
\appendix
\setcounter{table}{0}
\renewcommand{\thetable}{S\arabic{table}}
\setcounter{figure}{0}
\renewcommand{\thefigure}{S\arabic{figure}}

\section{Supplementary results}

\subsection{One-time initial alignment enables learning without weight transport}
\vspace{-0.1cm}

\subsubsection{Layer-wise weight alignment dynamics}

\begin{figure}[h!]
	\centering
	\includegraphics[width=\textwidth]{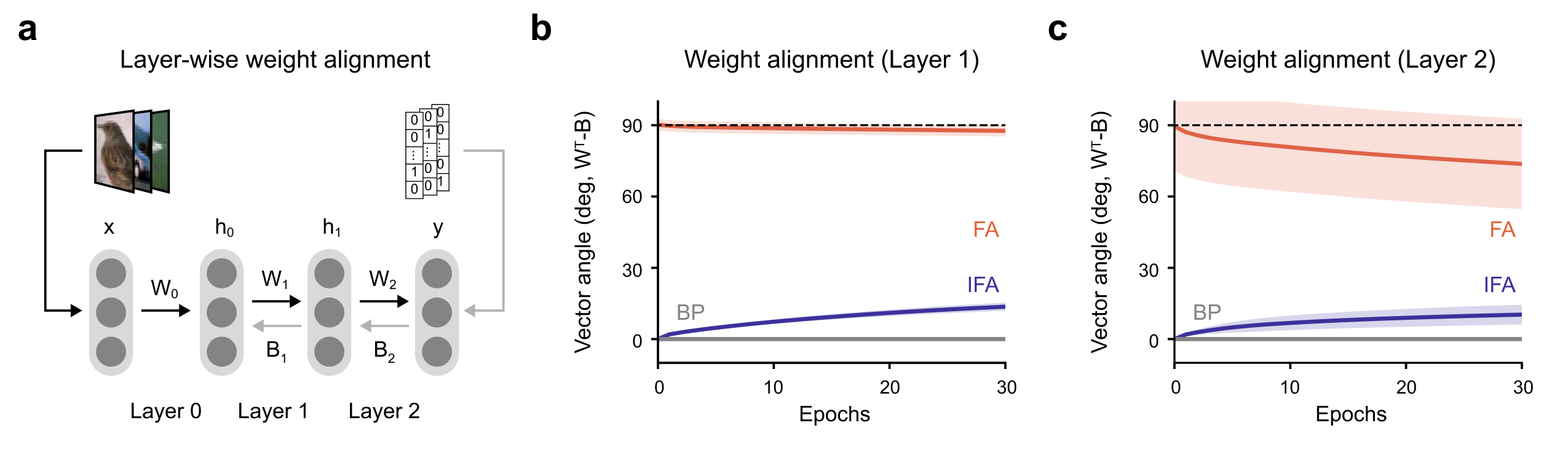}
    \vspace{-0.5cm}
    \caption{Weight alignment between forward and backward weights during learning.
        (a) Schematic of the three-layer feedforward neural network and its learning process. Forward and backward weights are analyzed in all layers except the first (Layer 0).
        (b-c) Alignment angle between forward and backward weights in each layer of the network.
        (b) Weight alignment in Layer 1. The shaded area represents the standard deviation of alignment angles across 512 neurons.
        (c) Weight alignment in Layer 2, the final layer of the network. The shaded area represents the standard deviation of alignment angles across 512 neurons.
    }
	\label{figs1}
\end{figure}

To generalize the trend of alignment dynamics observed in the penultimate layer (Figure 1f), we extended the analysis to all layers (Figure \ref{figs1}a). With the exception of the first layer, which lacks corresponding backward weights, we examined the vector angles between forward and backward weights in the second (Figure \ref{figs1}b) and final layers (Figure \ref{figs1}c). Consistent with the main results, feedback alignment (FA) \cite{lillicrap2016} learns by aligning forward weights with backward weights over the course of training, whereas standard backpropagation (BP) \cite{rumelhart1986} enforces identical forward and backward weights. Our model, initial feedback alignment (IFA), learns by gradually relaxing this alignment during training.

\subsubsection{Training details for different learning models}

\begin{figure}[h!]
	\centering
	\includegraphics[width=0.66\textwidth]{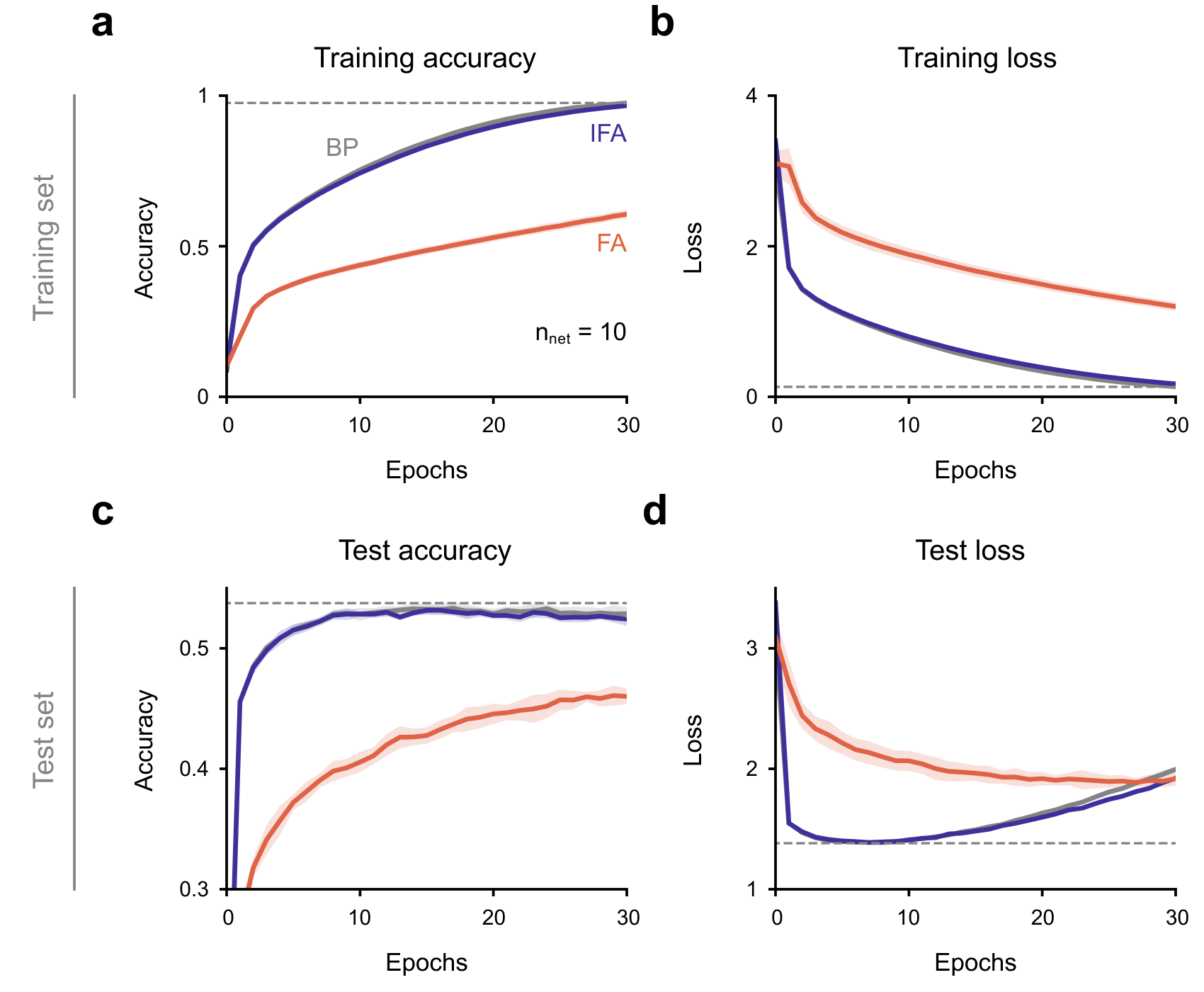}
    \vspace{-0.3cm}
    \caption{Learning curves across different feedback dynamics.
        (a–d) Learning curves over the course of training.
        (a) Training accuracy.
        (b) Training loss.
        (c) Test accuracy.
        (d) Test loss. The dotted gray horizontal line represents the final performance achieved by BP. The shaded area denotes the standard deviation across ten independent trials.
    }
	\label{figs2}
\end{figure}

To investigate the impact of initial alignment on neural network learning, we visualized learning curves under three different learning rules: (1) BP, (2) FA, and (3) IFA, while keeping all other experimental conditions constant. We plotted training accuracy (Figure \ref{figs2}a), training loss (Figure \ref{figs2}b), test accuracy (Figure \ref{figs2}c), and test loss (Figure \ref{figs2}d) over the course of training. Baseline FA exhibited slower convergence and lower performance compared to standard BP. In contrast, IFA achieved comparable learning speed and accuracy to BP, with nearly overlapping learning curves for both the training and test sets.

\subsubsection{Validation across neural network architectures}

\begin{figure}[h!]
	\centering
	\includegraphics[width=\textwidth]{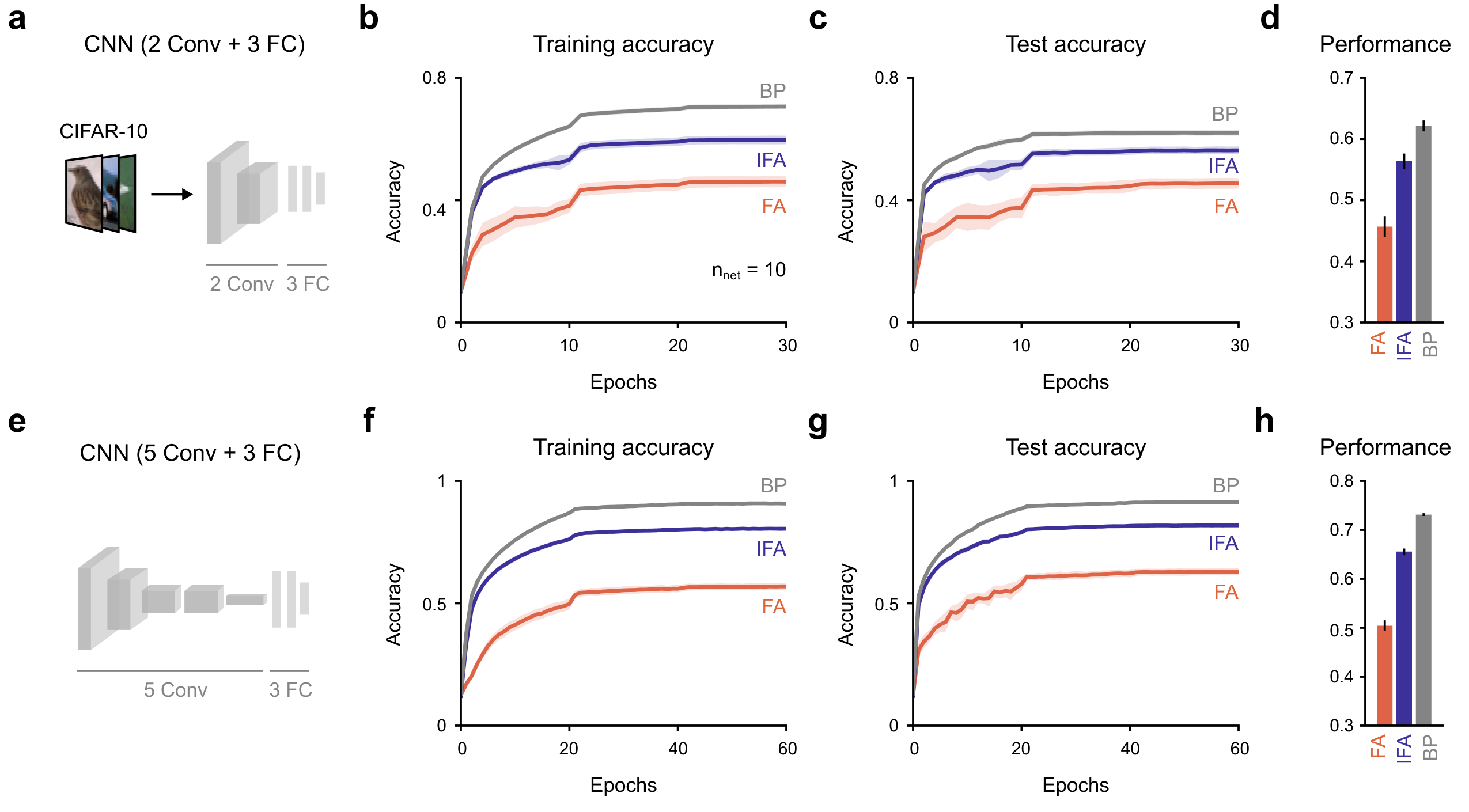}
    \caption{Effect of initial weight alignment on convolutional neural networks.
        (a) A convolutional neural network variant of LeNet, consisting of two convolutional layers and three fully connected layers, is used. The model is trained on CIFAR-10.
        (b) Training accuracy.
        (c) Test accuracy.
        (d) Final test accuracy.
        (e) A convolutional neural network variant of AlexNet, consisting of five convolutional layers and three fully connected layers, is used.
        (f) Training accuracy.
        (g) Test accuracy.
        (h) Final test accuracy. The shaded areas in the line plots and the error bars in the bar plots represent the standard deviation across ten independent trials.
    }
	\label{figs3}
\end{figure}

\begin{table}[h!]
\footnotesize
\caption{Benchmarked performance of the convolutional neural networks. Each performance value (\%) is reported as the mean~$\pm$~standard deviation across ten independent trials.}
\label{tables1}
\centering \begin{tabular}{c c c c}
\toprule
{} & {} & 2 Conv + 3 FC & 5 Conv + 3 FC\\
\midrule
\multicolumn{2}{c}{BP} & $62.13 \pm 0.90$ & $73.10 \pm 0.29$\\
\midrule
\multirow{1}{*}{FA} & 90° & $45.67 \pm 1.71$ & $50.41 \pm 1.11$\\
\midrule
\multirow{1}{*}{IFA} & 0° & $56.37 \pm 1.24$ & $65.55 \pm 0.62$\\

\bottomrule
\end{tabular}
\end{table}

Our main results, obtained using multi-layer feedforward neural networks, demonstrated that initial alignment enhances the scalability of feedback alignment algorithms. To assess whether this effect generalizes to other architectures and deeper models, we further evaluated convolutional neural networks (CNNs) (Figure \ref{figs3}). We first tested a LeNet-variant CNN \cite{lecun1998} with two convolutional layers and three fully connected layers (Figure \ref{figs3}a). We found that both the training and test accuracy of IFA were significantly higher than those of baseline FA, and the performance gap between IFA and BP was reduced (Figure \ref{figs3}b, c). The final test performance of IFA substantially outperformed that of baseline FA (Figure \ref{figs3}d; Table \ref{tables1}). We also evaluated an AlexNet-variant CNN \cite{krizhevsky2012} consisting of five convolutional layers and three fully connected layers (Figure \ref{figs3}e), and observed a similar trend (Figure \ref{figs3}f–h; Table \ref{tables1}). For training, we used the Adam optimizer with learning rates of 0.001 and 0.0001, and applied weight decay with a decay coefficient of 0.0001 to prevent overfitting. These findings suggest that IFA is broadly applicable to neural architectures beyond multi-layer perceptrons and can scale effectively to modern deep learning models.

\subsubsection{Benchmarking model performance on standard datasets}

\begin{figure}[h!]
	\centering
	\includegraphics[width=\textwidth]{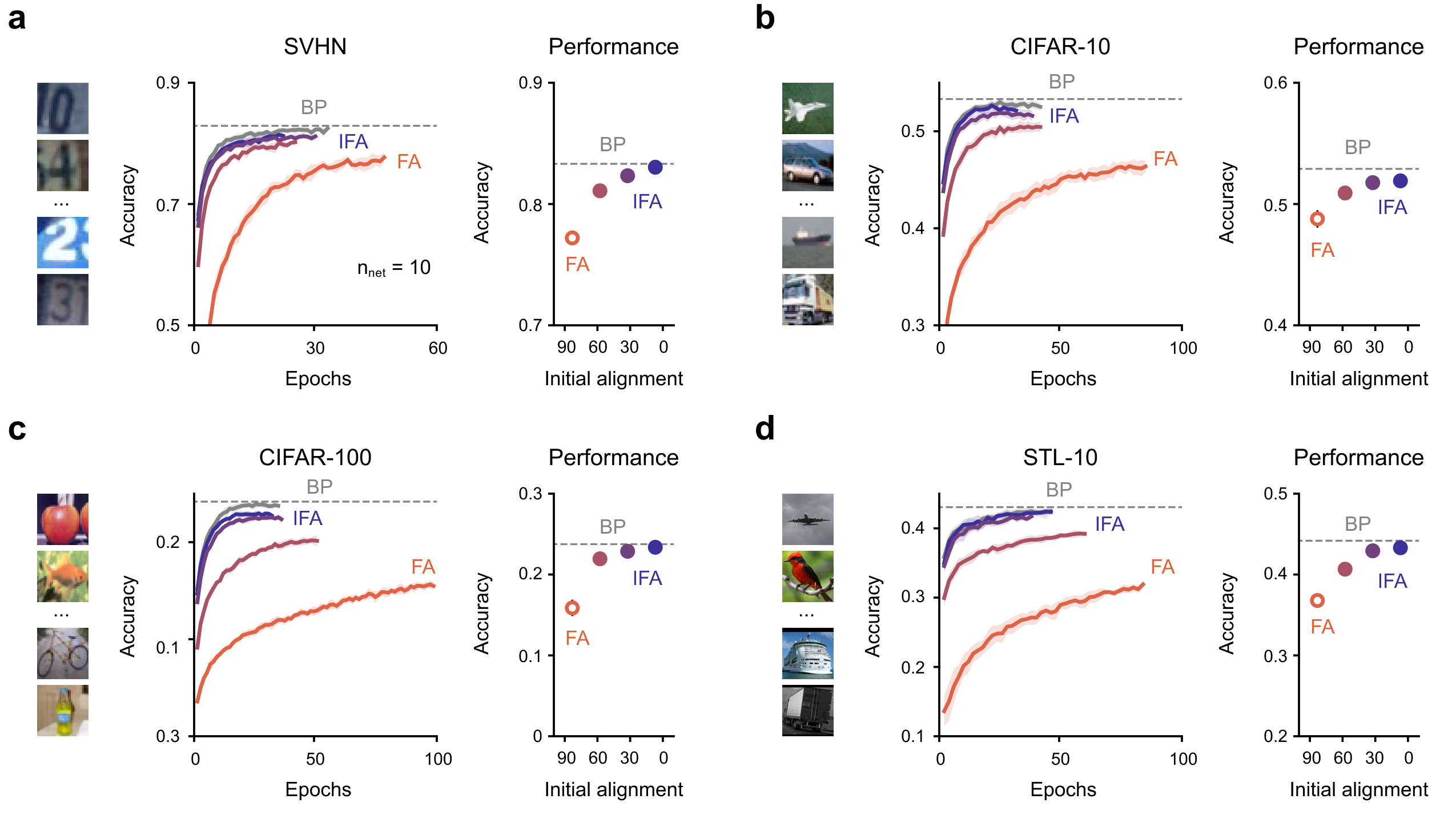}
    \caption{Benchmarked network performance across various image datasets.
        (a–d) Learning curves showing test accuracy over the course of training (left) and the benchmarked final test accuracy (right).
        (a) SVHN.
        (b) CIFAR-10.
        (c) CIFAR-100.
        (d) STL-10.
    }
	\label{figs4}
\end{figure}

\begin{table}[h!]
\footnotesize
\caption{Performance of the model on each dataset (SVHN, CIFAR-10, CIFAR-100, STL-10). Each performance value (\%) is reported as the mean~$\pm$~standard deviation across ten independent trials.}
\label{tables2}
\centering \begin{tabular}{c c c c c c}
\toprule
{} & {} & SVHN & CIFAR-10 & CIFAR-100 & STL-10 \\
\midrule
\multicolumn{2}{c}{BP} &       $82.90 \pm 0.11$ & $53.32 \pm 0.27$ & $24.18 \pm 0.28$ & $43.75 \pm 0.22$\\
\midrule
\multirow{1}{*}{FA}    & 90° & $78.78 \pm 0.72$ & $47.21 \pm 0.35$ & $16.79 \pm 0.57$ & $35.87 \pm 1.01$\\
\midrule
\multirow{4}{*}{IFA}   & 60° & $80.91 \pm 0.16$ & $51.09 \pm 0.43$ & $20.66 \pm 0.36$ & $41.95 \pm 0.24$\\
\cmidrule{2-6}
                       & 30° & $81.76 \pm 0.16$ & $52.33 \pm 0.30$ & $22.93 \pm 0.12$ & $42.87 \pm 0.30$ \\
\cmidrule{2-6}
                       & 0° & $81.91 \pm 0.20$ & $53.04 \pm 0.33$ & $23.31 \pm 0.20$ & $43.37 \pm 0.31$ \\

\bottomrule
\end{tabular}
\end{table}

Our main results were obtained using the CIFAR-10 dataset. To demonstrate that IFA generalizes across various datasets, we benchmarked the performance of BP, FA, and IFA on SVHN \cite{netzer2011} (Figure \ref{figs4}a), CIFAR-10 \cite{krizhevsky2009} (Figure \ref{figs4}b), CIFAR-100 \cite{krizhevsky2009} (Figure \ref{figs4}c), and STL-10 \cite{coates2011} (Figure \ref{figs4}d). Final accuracy was evaluated by extending training until test accuracy stopped improving for ten consecutive epochs (Table \ref{tables2}). Consistent with our main findings, the performance of baseline FA was significantly lower than that of BP. However, incorporating initial alignment substantially improved performance, in some cases reaching levels comparable to backpropagation. These results suggest that IFA is generalizable across different datasets and can effectively enhance learning performance.

\clearpage
\subsection{Stabilizing learning through initial weight alignment}

\subsubsection{Training networks across the parameter space}

\begin{figure}[h!]
	\centering
	\includegraphics[width=\textwidth]{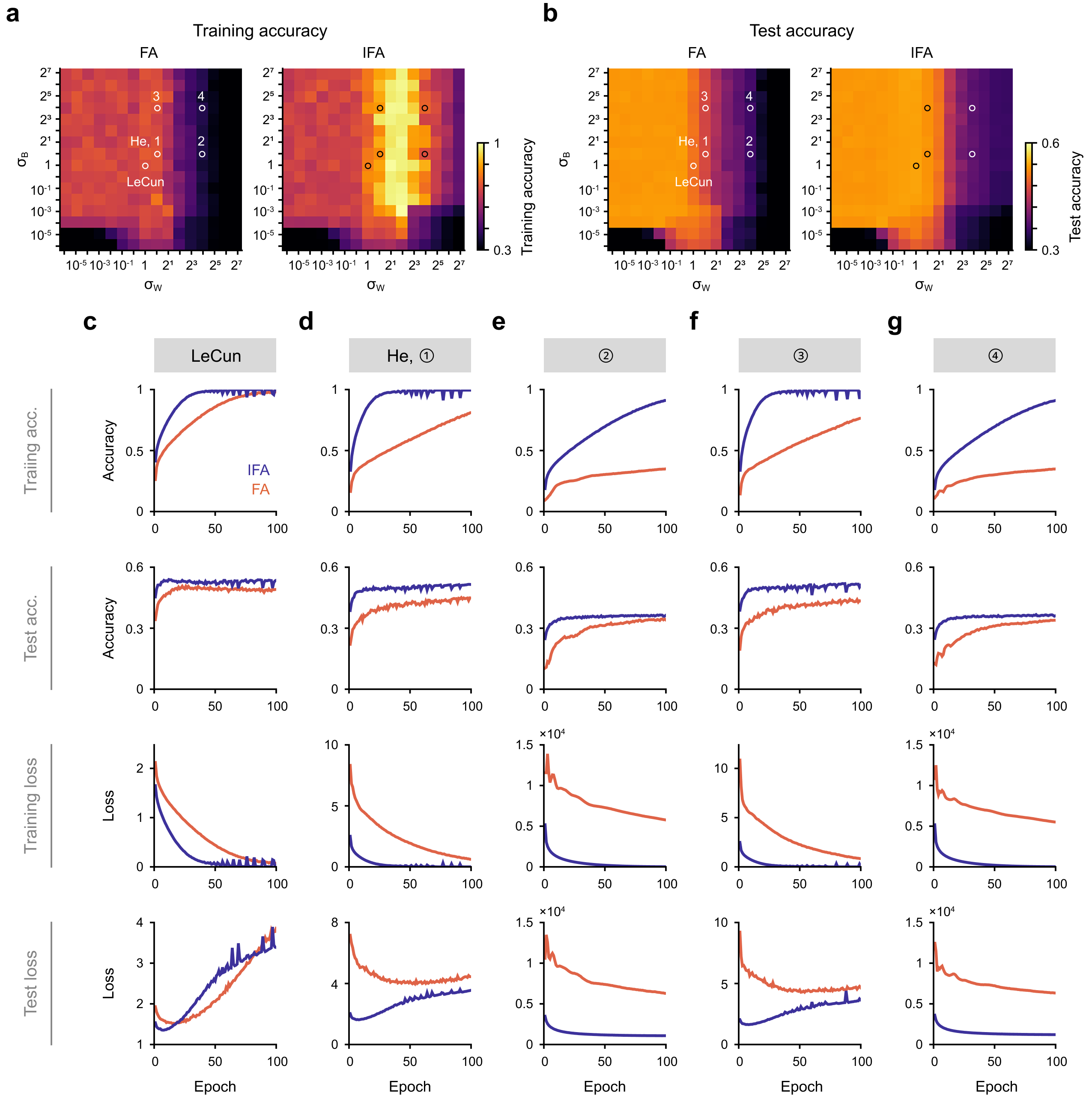}
    \caption{Trainability and generalization across the parameter space.
        (a) Training accuracy across the parameter space, with color indicating the final training accuracy.
        (b) Test accuracy across the parameter space, with color indicating the final test accuracy. The x- and y-axes represent the variances of the forward and backward weights, respectively. The point $(1, 1)$ corresponds to LeCun initialization (c), and $(\sqrt{2}, \sqrt{2})$ corresponds to He initialization (d; \textcircled{1}). Additionally, we investigate several representative conditions to visualize learning curves: large forward weight variance at $(2^4, \sqrt{2})$ (e; \textcircled{2}), large backward weight variance at $(\sqrt{2}, 2^4)$ (f; \textcircled{3}), and large variance in both weights at $(2^4, 2^4)$ (g; \textcircled{4}). For each variance condition, learning curves of training accuracy, test accuracy, training loss, and test loss are presented from top to bottom.
    }

	\label{figs5}
\end{figure}

In the main results, we reported the trainability of networks using FA and IFA across a parameter space defined by a two-dimensional grid of varying forward and backward weight variances (Figure 3a; Figure \ref{figs5}a). We observed that baseline FA supports only a narrow range of variance settings for achieving high training accuracy, whereas IFA enables a broader trainable parameter space. To extend this analysis, we also evaluated generalization by measuring test accuracy across the same parameter space (Figure \ref{figs5}b). Consistent with the trainability results, we found that initial alignment significantly expands the range of variance settings that yield high test accuracy. Furthermore, IFA consistently outperformed FA in nearly all regions of the parameter space.

To further characterize the learning behavior of FA and IFA, we selected five representative parameter settings and visualized the corresponding learning curves (Figure \ref{figs5}c–g). We first examined standard initialization methods: LeCun initialization (Figure \ref{figs5}c, $(1, 1)$) and He initialization (Figure \ref{figs5}d, $(\sqrt{2}, \sqrt{2})$, \textcircled{1}), confirming that IFA achieves faster convergence and higher performance in both training and test accuracy. We then extended the analysis to more extreme conditions, including large forward weight variance (Figure \ref{figs5}e, $(2^4, \sqrt{2})$, \textcircled{2}), large backward weight variance (Figure \ref{figs5}f, $(\sqrt{2}, 2^4)$, \textcircled{3}), and large variance in both forward and backward weights (Figure \ref{figs5}g, $(2^4, 2^4)$, \textcircled{4}). While FA exhibited extreme instability under these conditions, IFA demonstrated robust learning dynamics, with improved trainability and generalization.

\subsubsection{Training networks with varying depth}

\begin{figure}[h!]
	\centering
	\includegraphics[width=\textwidth]{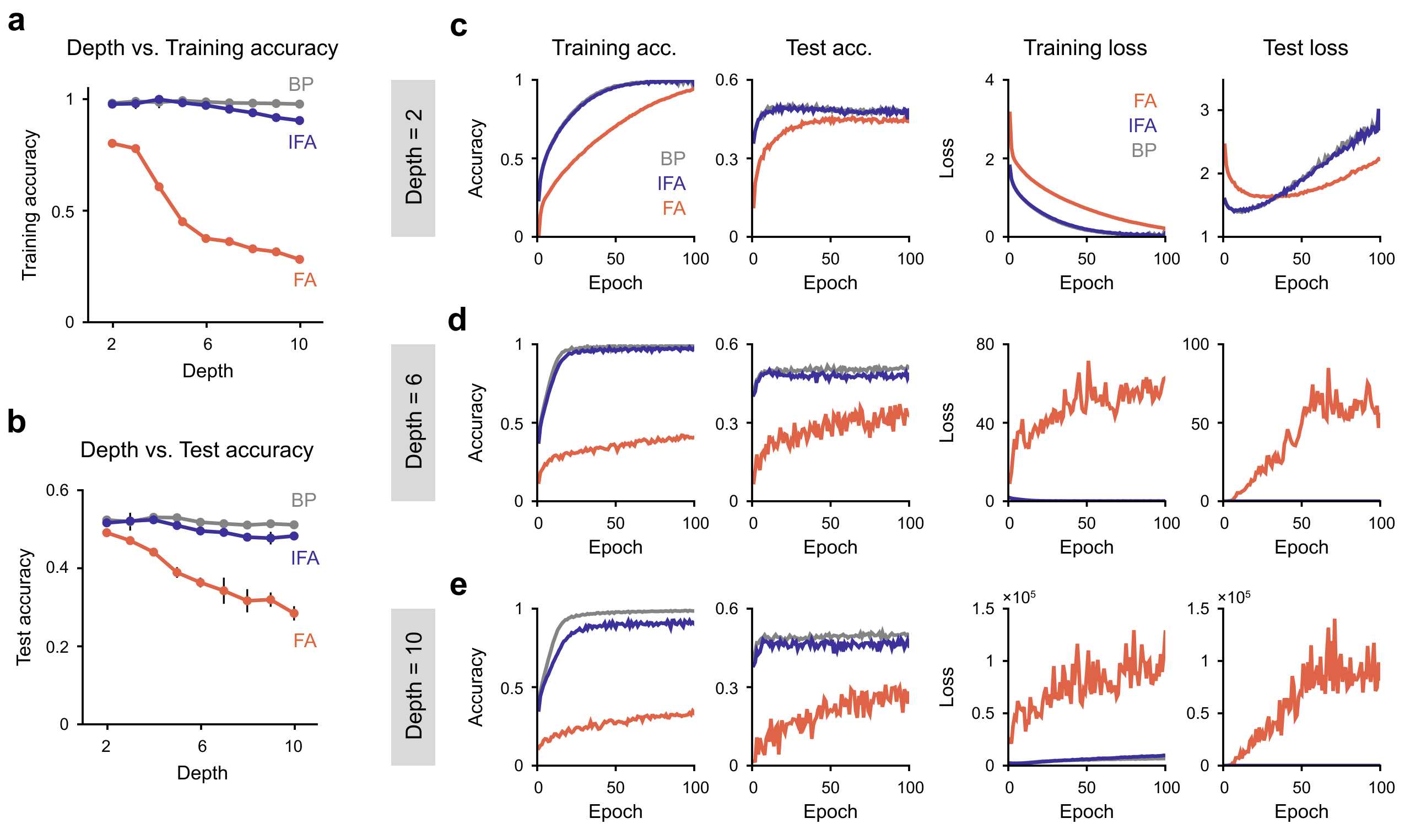}
    \caption{Trainability and generalization across different network depths.
        (a) Training accuracy as a function of network depth, for feedforward networks with depths ranging from 2 to 10 layers.
        (b) Test accuracy as a function of network depth, using the same range.
        (c-e) Learning curves for representative depths: 2 (c), 6 (d), and 10 (e). For each depth, training accuracy, test accuracy, training loss, and test loss are shown from left to right.
    }
	\label{figs6}
\end{figure}

We also extended our trainability analysis to networks with varying depths (Figure 3b; Figure \ref{figs6}a) and benchmarked test accuracy to assess generalization performance (Figure \ref{figs6}b). Consistent with the trainability results, IFA achieved significantly higher test accuracy than baseline FA and performed comparably to BP. Notably, FA exhibited a steep decline in test performance as network depth increased, whereas IFA maintained high test accuracy even in deeper architectures. To further investigate this trend, we selected three representative network depths and visualized the corresponding learning curves: depth 2 (Figure \ref{figs6}c), depth 6 (Figure \ref{figs6}d), and depth 10 (Figure \ref{figs6}e). As depth increased, baseline FA showed highly unstable learning dynamics — loss values diverged, and accuracy fluctuated significantly. In contrast, IFA demonstrated robust and stable learning, closely matching the behavior of standard BP. Interestingly, in deeper networks, FA’s loss continued to increase even as accuracy slightly improved. This discrepancy between loss and accuracy suggests a breakdown in model calibration: the network's confidence estimates become misaligned with actual prediction performance.

\subsubsection{Training with limited data}

\begin{figure}[h!]
	\centering
	\includegraphics[width=\textwidth]{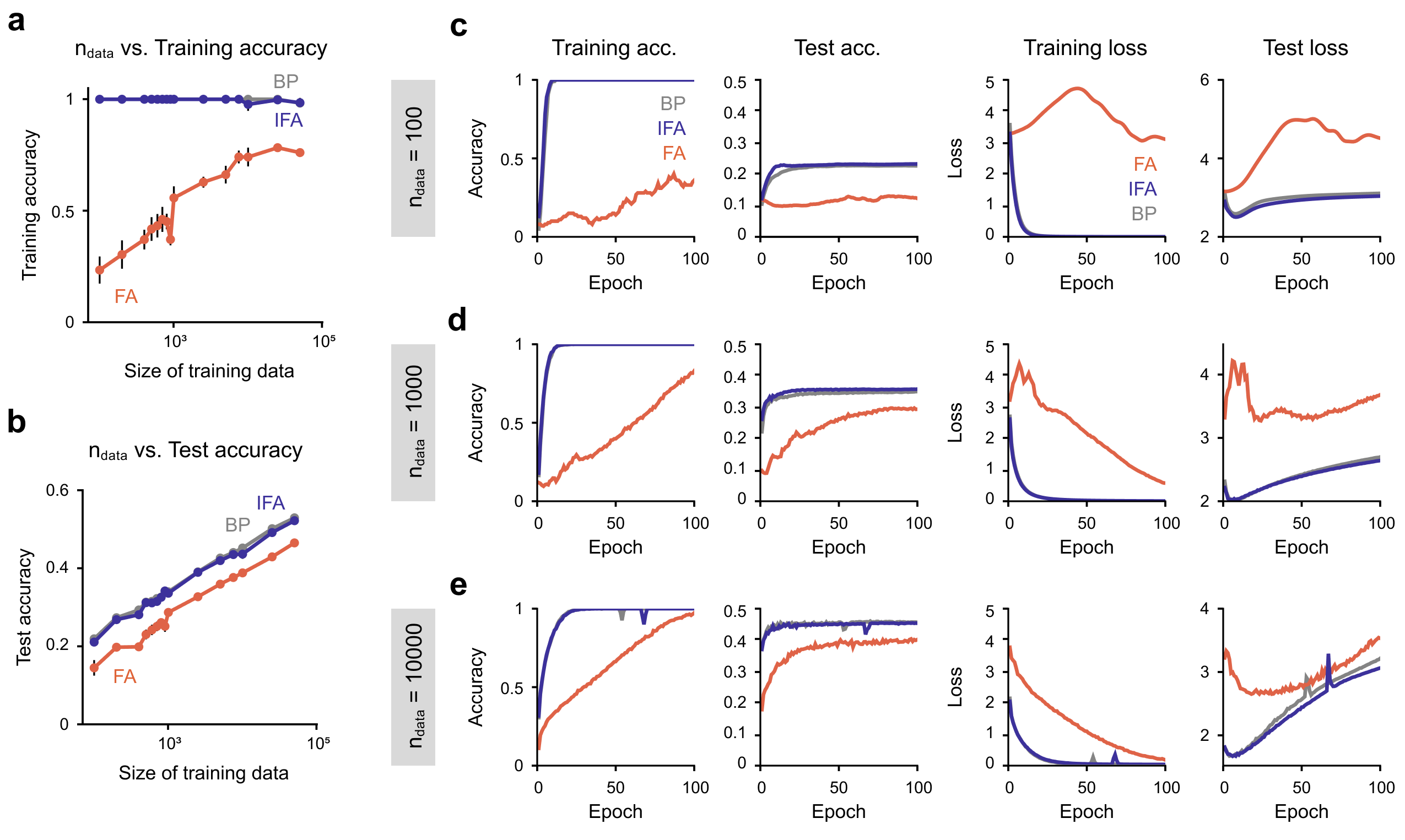}
    \caption{Trainability and generalization under limited data.
        (a) Training accuracy as a function of training data size, for feedforward networks trained on datasets ranging from 100 to 50,000 samples.
        (b) Test accuracy as a function of training data size, using the same range.
        (c-e) Learning curves for representative data sizes: 100 samples (c), 1,000 samples (d), and 10,000 samples (e). For each data size, training accuracy, test accuracy, training loss, and test loss are shown from left to right.
    }

	\label{figs7}
\end{figure}

Next, we investigated the effect of limited training data size (Figure 3c; Figure \ref{figs7}a) on generalization performance (Figure \ref{figs7}b). We benchmarked the test accuracy of networks trained on varying amounts of data. Notably, baseline FA consistently exhibited poor test performance compared to BP when trained on the same number of samples. In contrast, IFA demonstrated substantially improved generalization over FA, often achieving performance levels close to those of BP. The relationship between dataset size and test accuracy is frequently discussed in the context of neural scaling laws. Within this framework, initial alignment effectively shifts the scaling curve upward, enabling improved performance at every dataset size. To further examine this behavior, we selected three representative training data conditions and visualized the corresponding learning curves: 100 samples (Figure \ref{figs7}c), 1,000 samples (Figure \ref{figs7}d), and 10,000 samples (Figure \ref{figs7}e). Baseline FA notably struggled to learn under limited data, likely due to its reliance on data-driven alignment over time. In contrast, IFA produced stable and effective learning dynamics, even in data-scarce regimes.

\subsection{Initial alignment enhances robust generalization}

\subsubsection{Layer-wise dynamics of weight alignment}

\begin{figure}[t!]
	\centering
	\includegraphics[width=\textwidth]{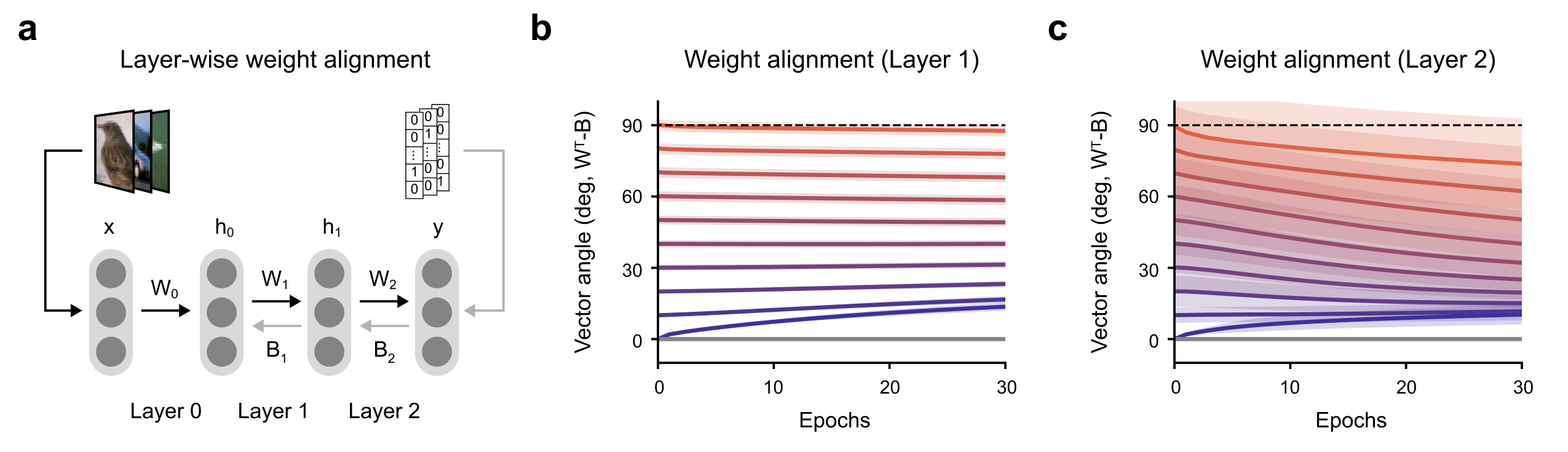}
    \caption{Effect of initial weight alignment on alignment dynamics during learning.
        (a) Schematic of a three-layer feedforward neural network. Forward and backward weights, excluding those in the first layer (Layer 0), are initialized with a soft alignment, ranging from $0^\circ$ to $90^\circ$. An initial angle of $90^\circ$ corresponds to the baseline feedback alignment.
        (b–c) Alignment angles between forward and backward weights in each layer.
        (b) Weight alignment in Layer 1. Shaded areas indicate the standard deviation of alignment angles across 512 neurons.
        (c) Weight alignment in Layer 2, the final layer of the network. Shaded areas indicate the standard deviation of alignment angles across 512 neurons.
    }
	\label{figs8}
\end{figure}

In the main results (Figure 4b), we visualized the alignment dynamics in the penultimate layer under various initial alignment conditions. We further extended this analysis to all layers (Figure \ref{figs8}a). With the exception of the first layer, we examined the angle between forward and backward weight vectors in both the second layer (Figure \ref{figs1}b) and the final layer (Figure \ref{figs1}c). Consistent with our main findings, the learning dynamics were highly dependent on the initial alignment. When the initial alignment was strong, the network gradually decreased the alignment during training. In contrast, when the initial alignment was weak, the network progressively increased the alignment through learning.

\subsubsection{Training details for different learning models}

\begin{figure}[h!]
	\centering
	\includegraphics[width=0.66\textwidth]{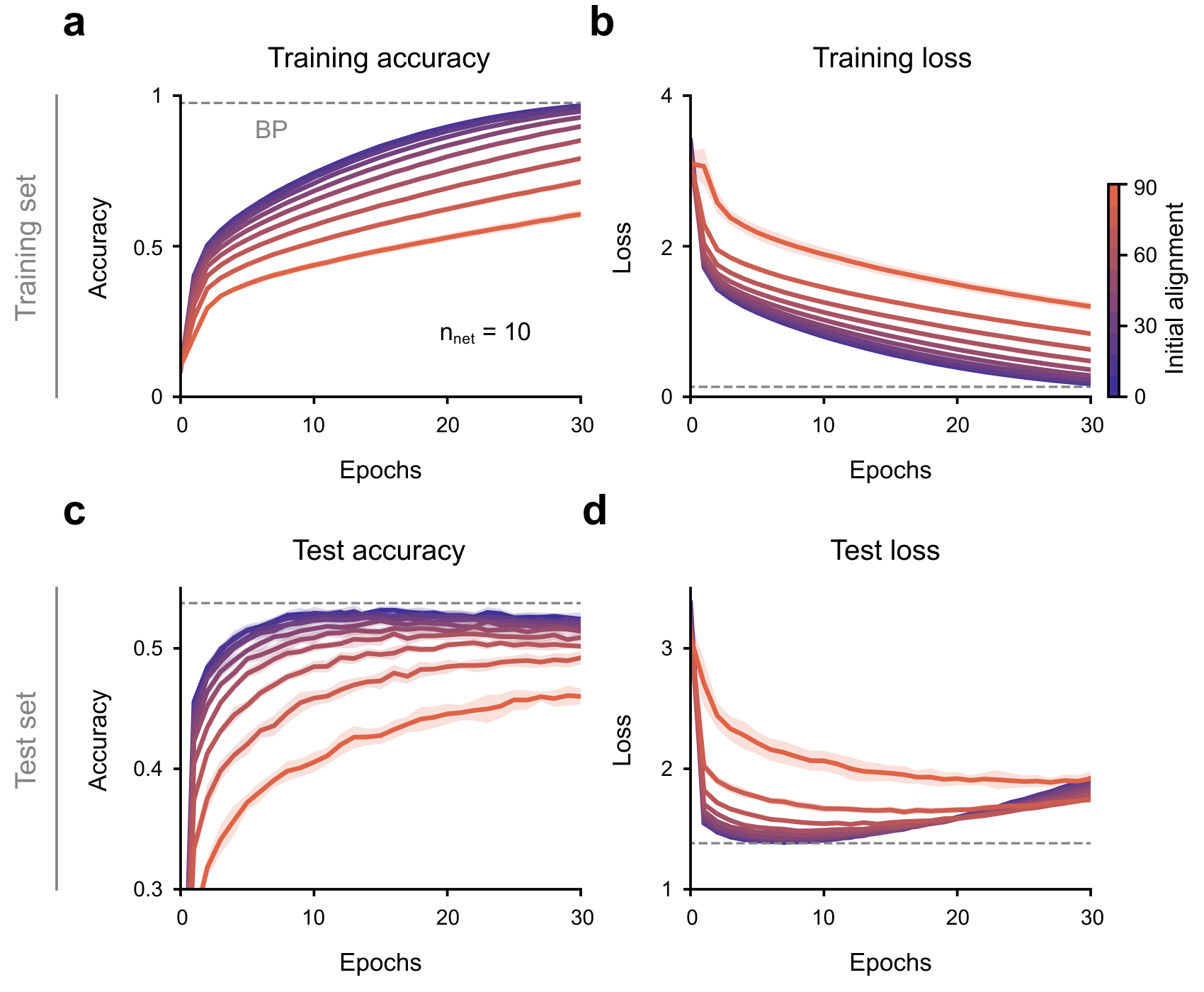}
    \caption{Learning curves across various initial alignment angles.
        (a–d) Learning curves over the course of training.
        (a) Training accuracy.
        (b) Training loss.
        (c) Test accuracy.
        (d) Test loss. The dotted gray horizontal line indicates the final performance of BP. Each solid line represents the learning curve of networks with a different initial alignment angle. Shaded areas indicate the standard deviation across ten independent trials.
    }
	\label{figs9}
\end{figure}

Next, we examined how initial alignment influences neural network learning. To isolate its effect, we visualized learning curves across a range of initial alignment angles while keeping all other experimental conditions constant. Specifically, we plotted training accuracy (Figure \ref{figs9}a), training loss (Figure \ref{figs9}b), test accuracy (Figure \ref{figs9}c), and test loss (Figure \ref{figs9}d) over the course of training. Our results show that although baseline FA performs poorly compared to BP, increasing the initial alignment markedly improves both learning speed and performance. As the alignment increases, the learning curves progressively resemble those of standard BP.

\subsubsection{Extended spectral analysis}

\begin{figure}[h!]
	\centering
	\includegraphics[width=\textwidth]{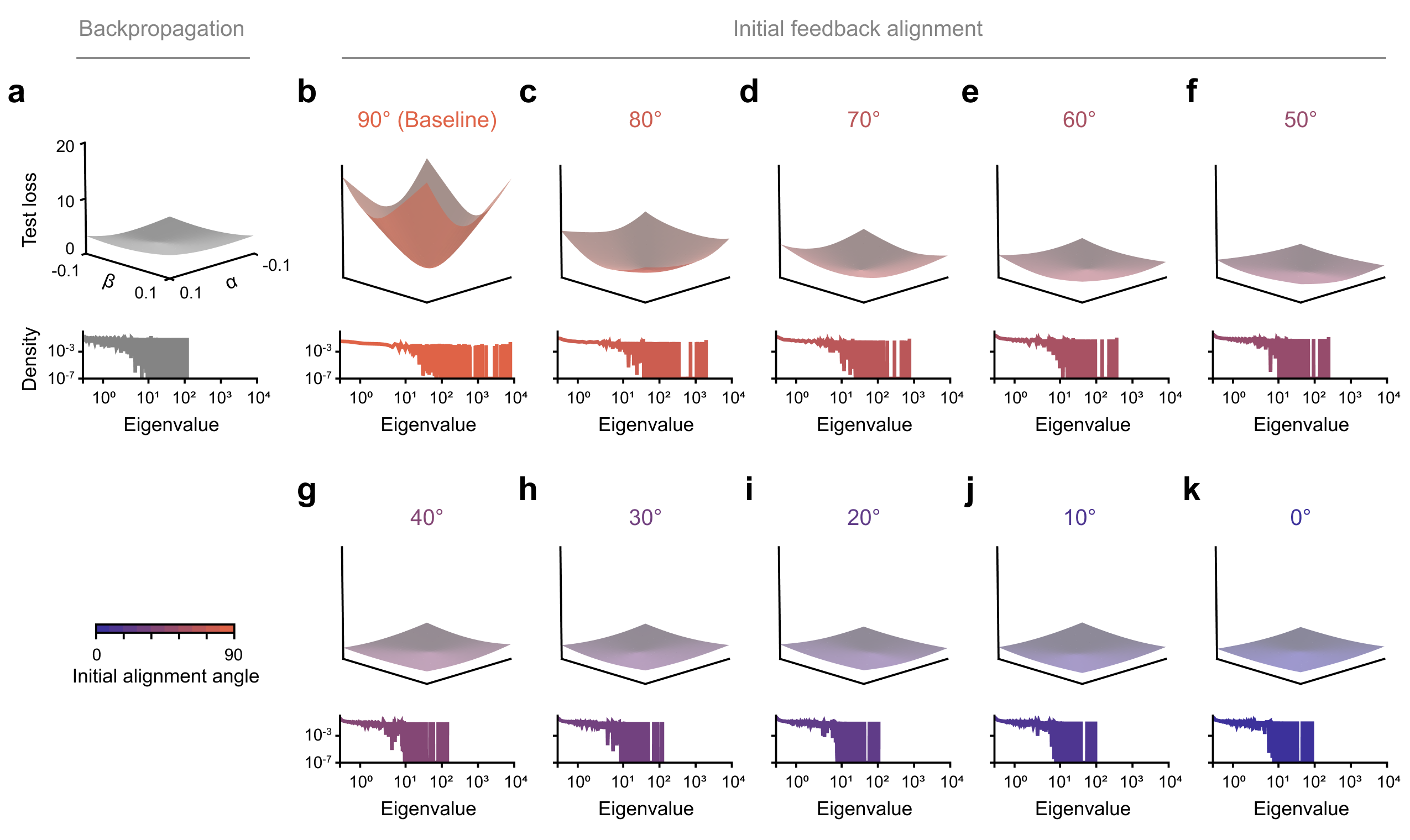}
    \vspace{-0.5cm}
    \caption{Spectral analysis of convergence behavior across initial alignment angles.
        (a–k) Loss landscapes visualized by perturbing the trained model parameters within a two-dimensional subspace defined by the top two eigenvectors of the Hessian (top). The axes, denoted by $\alpha$ and $\beta$, correspond to scaling factors for perturbations along the first and second eigenvector directions, respectively. Below each landscape, the spectral density of the Hessian eigenvalues is shown (bottom).
        (a) BP.
        (b–k) IFA with varying initial alignment angles ranging from $0^\circ$ to $90^\circ$. An initial angle of $90^\circ$ (b) corresponds to the baseline feedback alignment.
    }
	\label{figs10}
\end{figure}

To further investigate how initial weight alignment influences the convergence behavior of neural networks, we conducted a spectral analysis of the loss landscape (Figure \ref{figs10}). For each trained model, we visualized the local geometry of the loss surface by perturbing the parameters within a two-dimensional subspace spanned by the top two eigenvectors of the Hessian matrix. Specifically, we plotted the loss as a function of perturbations along these principal curvature directions, denoted by $\alpha$ and $\beta$, to assess the flatness and sharpness of the local minima. In addition, we examined the Hessian spectrum by plotting the eigenvalue density distribution. We first observed that BP exhibits a relatively flat loss landscape and a narrow eigenvalue spectrum (Figure \ref{figs10}a), in contrast to baseline FA, which shows a sharper landscape and a broader spectrum (Figure \ref{figs10}b). However, as the initial alignment improved (i.e., as the alignment angle decreased), the loss landscape became progressively flatter, and the Hessian spectra became increasingly concentrated within a narrower range (Figures \ref{figs10}c–k). These findings suggest that stronger initial alignment leads to smoother loss surfaces and more favorable curvature properties, ultimately enhancing robustness and generalization.

\subsubsection{Robustness under input corruption}

\begin{figure}[h!]
	\centering
	\includegraphics[width=\textwidth]{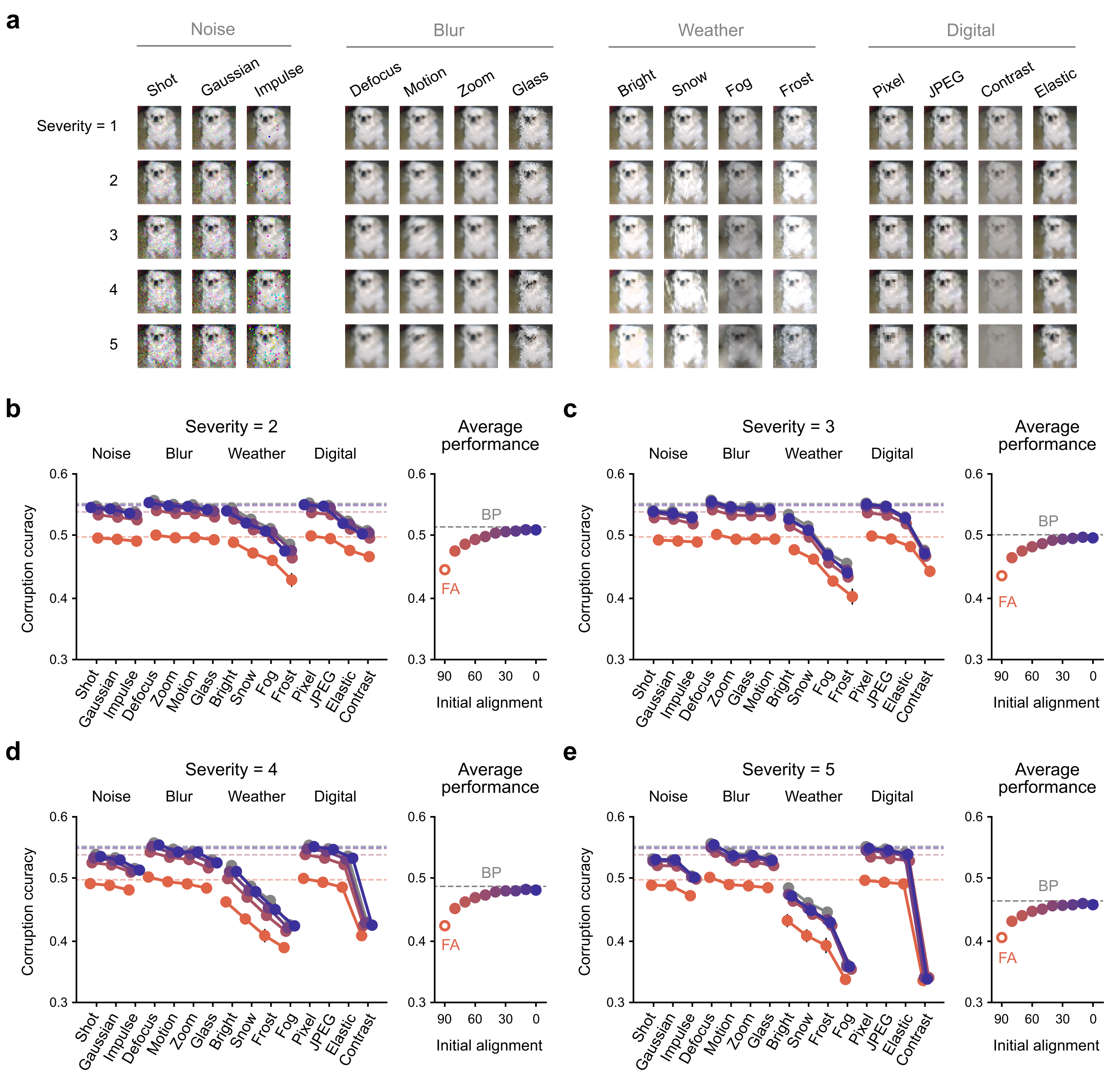}
    \vspace{-0.5cm}
    \caption{Effect of initial weight alignment on robustness to input corruption of varying severity.
        (a) Fifteen image corruption types, grouped into four categories—noise, blur, weather, and digital—were applied to test samples to assess network robustness. Each corruption type was tested across five severity levels, ranging from 1 (least severe) to 5 (most severe).
        (b–f) Classification accuracy on corrupted inputs (solid lines with markers) compared to accuracy on clean, uncorrupted images (dotted lines) (left). The relationship between initial alignment angle and robustness, measured as the average accuracy across all corruption types, is shown on the right.
        (b) Severity 2 (Severity 1 is shown in the main results).
        (c) Severity 3.
        (d) Severity 4.
        (e) Severity 5.
    }
	\label{figs11}
\end{figure}

To further assess the impact of initial weight alignment on model robustness, we systematically evaluated performance under varying levels of input corruption (Figure \ref{figs11}). Following a standard image corruption benchmark \cite{hendrycks2018}, we applied fifteen corruption types—categorized into noise, blur, weather, and digital—to the test set, each at five severity levels ranging from level 1 (least severe) to level 5 (most severe) (Figure \ref{figs11}a). While results for severity level 1 were presented in the main text (Figure 6b, c), here we report extended results for severity levels 2 through 5 (Figures \ref{figs11}b–e). We compared the corruption robustness of networks initialized with different alignment conditions by measuring their classification accuracy on both clean and corrupted inputs. Consistent with our main results, FA showed substantially lower robustness compared to standard BP. However, increasing the degree of initial alignment significantly improved robustness, bringing performance closer to that of BP. Although all models exhibited performance degradation as corruption severity increased, the decline was significantly less pronounced in networks with stronger initial alignment. These results highlight the robustness benefits conferred by initial alignment—networks with better-aligned initial weights demonstrated greater resilience to input corruptions, suggesting that alignment at initialization helps stabilize learning and supports more reliable performance under distributional shifts. This observation aligns with our earlier spectral analysis of the loss landscape, which showed that stronger initial alignment leads to flatter optima.

\vspace{-0.2cm}
\subsection{The role of weight misalignment in promoting adversarial robustness}

\subsubsection{Adversarial robustness to Basic Iterative Method (BIM)}

\begin{figure}[t!]
	\centering
	\includegraphics[width=0.66\textwidth]{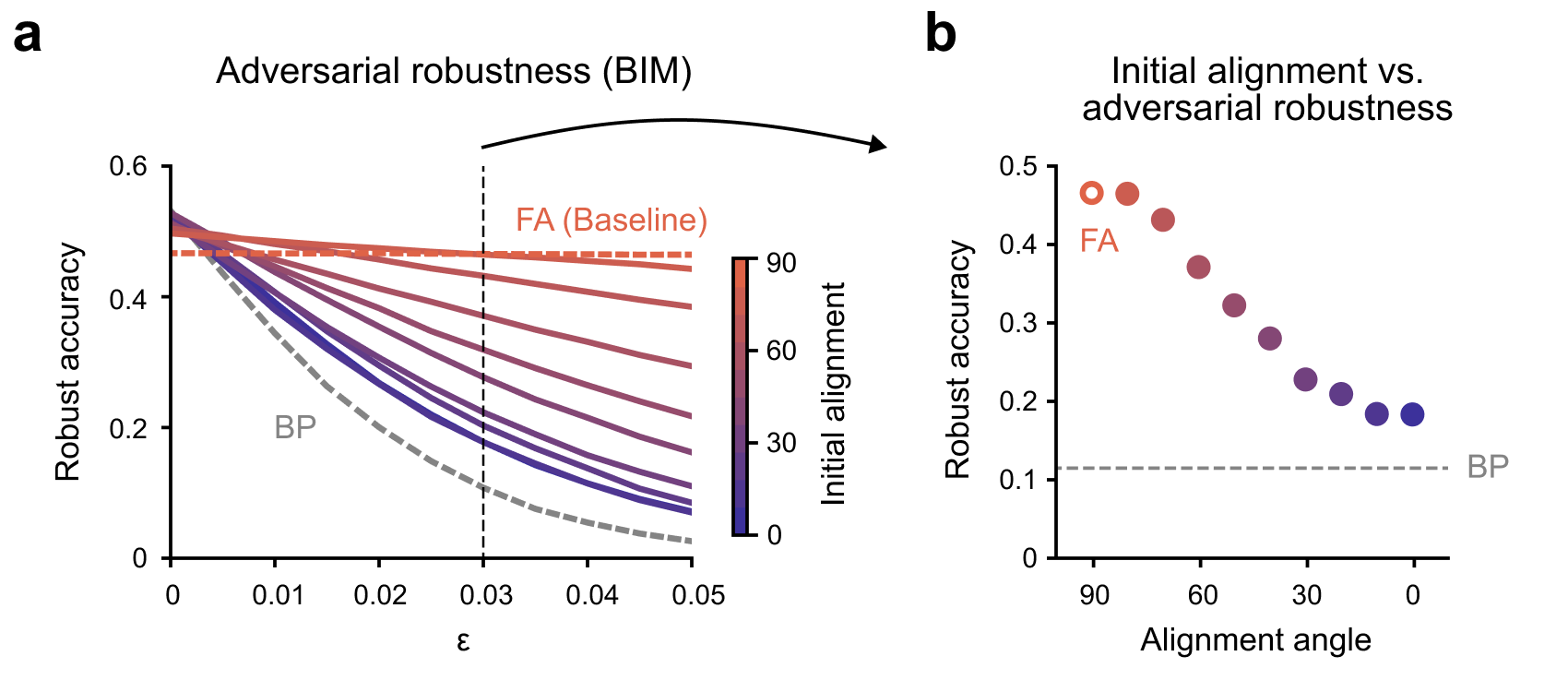}
    \caption{Effect of weight misalignment on adversarial robustness to the Basic Iterative Method (BIM).
        (a) Classification accuracy as a function of increasing adversarial perturbation strength under attack using the BIM.
        (b) Relationship between the initial alignment angle and adversarial robustness, measured as classification accuracy under attack with perturbation magnitude $\epsilon = 0.03$.
    }
	\label{figs12}
\end{figure}

In the main results, we demonstrated that the misalignment introduced by IFA enhances robustness against adversarial attacks, particularly using the Fast Gradient Sign Method (FGSM) (Figure 7). To further support this finding and extend the robustness of IFA to other gradient-based attacks, we also evaluated performance under the Basic Iterative Method (BIM) \cite{kurakin2018}. BIM is an extension of FGSM that applies small, iterative perturbations in the direction of the gradient of the loss with respect to the input. At each step, the perturbation is constrained within a defined epsilon-ball around the original input to preserve perceptual similarity. We applied BIM to networks trained with various initial alignment angles and measured classification accuracy under increasing adversarial perturbation (Figure \ref{figs12}a). Consistent with our FGSM results, model accuracy under BP rapidly declined as perturbation strength increased, whereas models trained with IFA maintained significantly higher accuracy even under stronger attacks (Figure \ref{figs12}b).

\subsubsection{Adversarial robustness to Projected Gradient Descent (PGD)}

\begin{figure}[h!]
	\centering
	\includegraphics[width=0.66\textwidth]{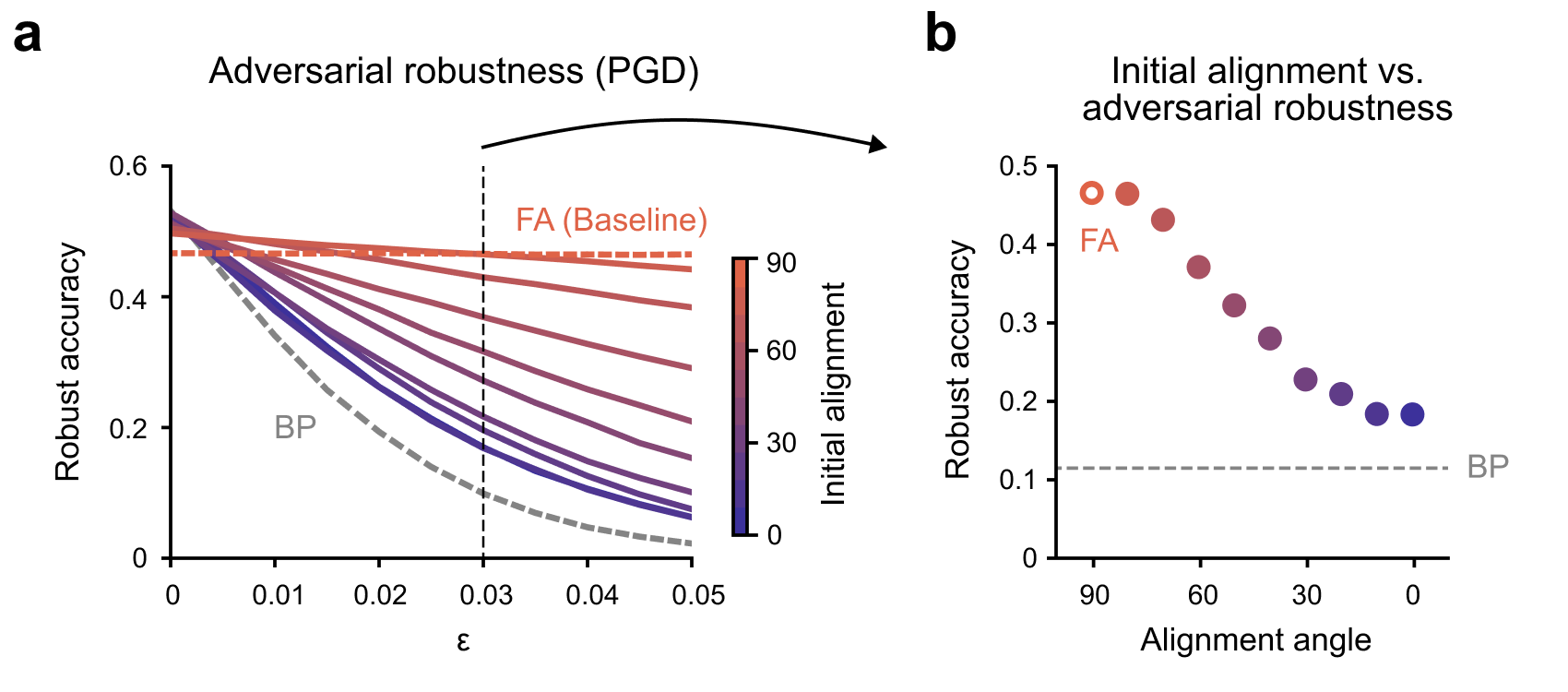}
    \caption{Effect of weight misalignment on adversarial robustness to the Projected Gradient Descent (PGD).
        (a) Classification accuracy as a function of increasing adversarial perturbation strength under attack using the PGD.
        (b) Relationship between the initial alignment angle and adversarial robustness, measured as classification accuracy under attack with perturbation magnitude $\epsilon = 0.03$.
    }
	\label{figs13}
\end{figure}

We also evaluated robustness against the Projected Gradient Descent (PGD) attack \cite{madry2017}, widely considered one of the most effective first-order adversarial attacks. PGD is an iterative extension of the Basic Iterative Method (BIM) that incorporates a projection step to ensure that adversarial examples remain within an $\epsilon$-bounded $\ell_\infty$-ball around the original input. At each iteration, the input is updated by ascending the gradient of the loss with respect to the input, followed by a projection back into the allowed perturbation region if the update exceeds the specified bound. We applied PGD to networks with varying initial alignment angles and measured their accuracy under attack (Figure \ref{figs13}a). Consistent with the results from FGSM and BIM, IFA-trained models showed significantly greater resilience to PGD compared to those trained with standard BP (Figure \ref{figs13}b). These findings reinforce the role of initial alignment in enhancing both generalization and robustness to strong adversarial attacks. In particular, they suggest that greater initial misalignment further improves adversarial robustness, highlighting a key functional advantage of IFA over standard BP.

\clearpage
\section{Methods}

\subsection{Neural network models}

We employed a multi-layer feedforward neural network $f_\theta$ for image classification throughout the research. Specifically, the network begins with an input layer that receives a flattened one-dimensional vector of the input image. It then processes the input through multiple hidden, fully connected layers with forward weights $\mathbf{W}_l \in \mathbb{R}^{m \times n}$ and biases $\mathbf{b}_l \in \mathbb{R}^{n}$. Each hidden layer applies a nonlinear activation function to its output. We used two hidden layers, each with 512 units, and employed the Rectified Linear Unit (ReLU) as the activation function. The final layer applies a SoftMax function to produce a probability distribution over image categories. The forward weights $\mathbf{W}_l$ were initialized using a normal distribution with zero mean and a standard deviation scaled according to the input size. Unlike standard backpropagation (BP), which uses the transpose of the forward weights for error propagation, feedback alignment (FA) utilizes separate backward weights $\mathbf{B}_l \in \mathbb{R}^{n \times m}$. These backward weights $\mathbf{B}_l$ are also initialized randomly from the same distribution as the corresponding forward weights but remain fixed during training.

\subsection{Soft Initial Alignment}

In baseline Feedback Alignment (FA), the forward weights $\mathbf{W}_l$ and backward weights $\mathbf{B}_l$ are initialized independently using random distributions. In contrast, our proposed method—initial feedback alignment (IFA)—introduces a soft alignment at initialization, wherein the forward weights are sampled from a subspace partially aligned with the backward weights. Specifically, the forward weights $\mathbf{W}_l$ are initialized by projecting onto a subspace spanned by the backward weights $\mathbf{B}_l^\top$ and an independent random matrix $\mathbf{R}$:

$$
\mathbf{B}_l \sim \mathcal{N}\left(0, \frac{a^2}{fan_{\text{in}}}\right), \quad \mathbf{R} \sim \mathcal{N}\left(0, \frac{a^2}{fan_{\text{in}}}\right)
$$
$$
\mathbf{W}_l = \mathbf{B}_l^\top \cos(\theta_{\text{init}}) + \mathbf{R} \sin(\theta_{\text{init}}),
$$

where $fan_{\text{in}}$ denotes the number of input units to the layer, and $a = \sqrt{2}$ follows the He initialization scheme, which is suitable for networks using ReLU activations. This initialization forms a linear combination of aligned and random components, with the parameter $\theta_{\text{init}}$ controlling the degree of initial alignment. When $\theta_{\text{init}} = 0^\circ$, the forward and backward weights are fully aligned ($\mathbf{W}_l = \mathbf{B}_l^\top$), whereas $\theta_{\text{init}} = 90^\circ$ corresponds to the standard FA case, where $\mathbf{W}_l$ and $\mathbf{B}_l^\top$ are orthogonal and thus unaligned.

\subsection{Neural networks training}

To evaluate the effects of different feedback dynamics, we trained neural networks using three distinct learning rules that differ only in how feedback signals are computed: (1) standard backpropagation (BP), (2) feedback alignment (FA), and (3) initial feedback alignment (IFA). Except for the feedback pathway, all other training conditions were held constant to ensure a fair comparison across methods. We employed the Adam optimizer with hyperparameters $\beta_1 = 0.99$, $\beta_2 = 0.999$, and a fixed learning rate of $10^{-4}$. The training was performed using a batch size of 128 for 30 epochs in the main experiments, with extended training conducted for benchmarking or additional analyses when necessary. The primary dataset used was CIFAR-10, which comprises 50,000 training images and 10,000 test images. To assess generalizability, we also evaluated our methods on the CIFAR-100 and STL-10 datasets; these results are reported in the Supplementary Results.

\subsection{Measurement of the alignment between forward and backward weights}

To quantitatively assess the alignment between the forward weights $\mathbf{W}_l \in \mathbb{R}^{m \times n}$ and the backward weights $\mathbf{B}_l \in \mathbb{R}^{n \times m}$ at layer $l$, we computed the angular similarity between corresponding forward and backward weight vectors. For the $i$-th neuron in layer $l$, the forward weight vector is defined as the $i$-th row of $\mathbf{W}_l^\top$, denoted $\mathbf{W}_l^{^\top(i)} \in \mathbb{R}^n$, and the corresponding backward weight vector is the $i$-th row of $\mathbf{B}_l$, denoted $\mathbf{B}_l^{(i)} \in \mathbb{R}^n$. The alignment angle between the two vectors was then computed using the cosine similarity \cite{lillicrap2016}:

$$
\theta_i = \cos^{-1} \left( \frac{ \mathbf{W}_l^{^\top(i)} \cdot \mathbf{B}_l^{(i)} }{ \| \mathbf{W}_l^{^\top(i)} \| \, \| \mathbf{B}_l^{(i)} \| } \right),
$$

where $\theta_i$ represents the alignment angle for the $i$-th neuron. A smaller angle (closer to $0^\circ$) indicates stronger alignment between the forward and backward weights, whereas a larger angle (closer to $90^\circ$) indicates weaker or no alignment. We reported the average alignment across all neurons as a summary metric of the overall correspondence between the forward and backward pathways in the network.

\subsection{Trainability analysis}

To evaluate whether neural networks remain trainable under different feedback dynamics, we assessed trainability by benchmarking accuracy with various conditions. Specifically, we trained neural network models for 100 epochs and recorded the maximum training accuracy achieved. We examined the effects of several conditions on trainability: (1) variance of forward and backward weights, (2) network depth, and (3) size of the training dataset. While our primary focus was on training accuracy as a measure of trainability, we also reported test accuracy in the Supplementary Results.

First, we evaluated the trainable parameter space by systematically varying the variances of the forward and backward weights. Networks were initialized using zero-mean normal distributions with different variances:

\[
\mathbf{W}_l \sim \mathcal{N}\left(0, \frac{a^2}{fan_{\text{in}}}\right), \quad \mathbf{B}_l \sim \mathcal{N}\left(0, \frac{b^2}{fan_{\text{in}}}\right),
\]

where $fan_{\text{in}}$ denotes the number of input units to each layer, and $a$ and $b$ are scaling parameters controlling the variance. Biases were initialized to zero. For example, $(a, b) = (1, 1)$ corresponds to LeCun initialization, and $(\sqrt{2}, \sqrt{2})$ corresponds to He initialization. We also explored a wide range of variances by varying $a$ and $b$ from $10^{-6}$ to $10^{0}$ (smaller variances), and from $2^{0}$ to $2^{7}$ (larger variances), using exponential step sizes of 1. Additional values of $10^{-0.5}$ and $2^{-0.5}$ were also included. This systematic investigation allowed us to visualize the trainable parameter space in a two-dimensional plot, where the x- and y-axes correspond to the variances of the forward and backward weights, respectively.

Next, we examined the effect of network depth on trainability. We trained multi-layer feedforward networks with depths ranging from 2 to 10, keeping the hidden layer size fixed at 512 units. All other conditions—including weight initialization, training dataset size, optimizer, and hyperparameters—were identical to those used in the main experiments.

Finally, we tested trainability under limited data conditions. We trained three-layer feedforward networks on subsets of the training data ranging from 100 to 50,000 samples. Again, all other training conditions were kept consistent with those used in the main experiments.

\subsection{Visualization of trajectory in loss landscape}

To visualize the optimization trajectory within the loss landscape \cite{li2018}, we recorded the model parameters $\theta_i$ at each training step and flattened them into one-dimensional vectors. We then constructed a matrix of difference vectors:

\[
[\theta_0 - \theta_{\text{final}}, \theta_1 - \theta_{\text{final}}, \theta_2 - \theta_{\text{final}}, \cdots],
\]

where each column represents the deviation of the parameter vector at a given step from the final converged model $\theta_{\text{final}}$. Principal Component Analysis (PCA) was applied to this matrix to identify the two principal directions that explain the most variance in the trajectory. These directions define a two-dimensional subspace within the high-dimensional parameter space. To visualize the loss surface, we interpolated model parameters across a grid within this subspace and computed the corresponding training loss values, thereby constructing a two-dimensional loss landscape. The optimization trajectory was projected onto the same subspace and overlaid on the loss surface, providing a geometric visualization of how the model navigates the loss landscape during training.

\subsection{Spectral analysis on Hessian of loss} 

To quantitatively analyze the stability of the learning trajectory and the characteristics of converged optima, we investigated the geometric properties of the loss landscape by performing spectral analysis using the Hessian of the loss. Since we mainly considered neural networks for image classification tasks, we define the network as $f_\theta$, parameterized by $\theta = \{\mathbf{W}_l, \mathbf{b}_l\}_{l=0}^{L-1}$, and optimized to minimize the loss function $\mathcal{L}(\theta)$. The gradient and second derivative of the loss with respect to model parameters, i.e., the Hessian, are given by:

$$
g_\theta = \frac{\partial \mathcal{L}}{\partial \theta}, \quad
H_\theta = \frac{\partial^2 \mathcal{L}}{\partial \theta^2} = \frac{\partial g_\theta}{\partial \theta}.
$$

However, due to the large number of parameters in deep neural networks, explicitly computing the full Hessian matrix is computationally infeasible. Therefore, we analyzed the spectral properties of the Hessian without explicitly forming it, using approximate and iterative methods as proposed in previous research \cite{yao2020}. In particular, this can be achieved by computing the product of the Hessian and a random probe vector $v \sim \mathcal{N}(0, I)$. This product, $H_\theta v$, can be efficiently computed by backpropagating the inner product $g_\theta^\top v$, using the identity:

$$
\frac{\partial g_\theta^\top v}{\partial \theta} = \frac{\partial g_\theta^\top}{\partial \theta} v + g_\theta^\top \frac{\partial v}{\partial \theta} = \frac{\partial g_\theta^\top}{\partial \theta} v = H_\theta v.
$$

We used power iteration to compute the top eigenvalue of the Hessian \cite{yao2018}. Next, we computed the trace of the Hessian using a randomized numerical linear algebra method \cite{avron2011}. The trace is computed using the identity:

$$
\mathrm{Tr}(H_\theta) = \mathrm{Tr}(H_\theta I) = \mathrm{Tr}(H_\theta \mathbb{E}[vv^\top]) = \mathbb{E}[\mathrm{Tr}(H_\theta vv^\top)] = \mathbb{E}[v^\top H_\theta v],
$$

where $v$ is a random probe vector. Since $v^\top H_\theta v$ is the dot product between $H_\theta v$ and $v$, the trace can be estimated by averaging this quantity over multiple samples. We also estimated the empirical spectral density (ESD) of the Hessian using the stochastic Lanczos quadrature (SLQ) method \cite{golub1969, ghorbani2019}. The ESD is defined as

$$
\rho(\lambda) = \frac{1}{m} \sum_{i=1}^m \delta(\lambda - \lambda_i),
$$

where $\lambda_i$ are the eigenvalues of the Hessian matrix. Since computing all eigenvalues is intractable, we approximate $\rho(\lambda)$ by convolving it with a Gaussian kernel, resulting in a smoothed estimate of the density. To compute this efficiently, we apply the Lanczos algorithm to Hessian-vector products using multiple random probe vectors. This yields a tridiagonal matrix whose eigenvalues and weights approximate those of the original Hessian. The final estimate of the smoothed spectral density is given by:

$$
\rho_\sigma(t) \approx \frac{1}{n_v} \sum_{l=1}^{n_v} \sum_{i=1}^q \tau_i^{(l)} \cdot \frac{1}{\sigma \sqrt{2\pi}} \exp\left(-\frac{(t - \lambda_i^{(l)})^2}{2\sigma^2}\right),
$$

where $\lambda_i^{(l)}$ and $\tau_i^{(l)}$ are the eigenvalues and associated weights from the $l$-th Lanczos run, and $n_v$ is the number of random probe vectors used. Throughout our spectral analysis, we utilized and modified code provided by previous research \cite{yao2020}.

\subsection{Perturbed loss landscape analysis}

To evaluate the stability and smoothness of the converged solution, we analyzed the perturbed loss landscape around the final model parameters. After convergence, we computed the top two eigenvectors of the Hessian of the loss function, corresponding to the largest eigenvalues. These eigenvectors define the principal directions of curvature in the parameter space. We then constructed a two-dimensional subspace by perturbing the converged parameters along these directions. Specifically, for scaling parameters $\alpha$ and $\beta$, the perturbed parameters were defined as:

\[
\theta_{\text{perturbed}} = \theta_{\text{final}} + \alpha \lambda_1 + \beta \lambda_2,
\]

where $\lambda_1$ and $\lambda_2$ are the eigenvectors associated with the largest and second-largest eigenvalues, respectively. The point $(\alpha, \beta) = (0, 0)$ corresponds to the unperturbed test loss at the converged solution, while losses evaluated at other points reflect the model’s sensitivity to perturbations in parameter space. A relatively flat loss landscape—where test loss remains low even under substantial perturbations—indicates that the solution resides in a wide, stable minimum. Conversely, a sharp increase in loss with small perturbations suggests that the solution lies in a narrow or sharp minimum, which may be associated with poorer generalization and greater sensitivity to parameter variations.

\subsection{Corruption robustness}

To assess the model’s robustness against input corruptions, we evaluated its performance on the CIFAR-10-C dataset \cite{hendrycks2018} after training on the clean CIFAR-10 dataset. CIFAR-10-C includes 15 types of corruptions that mimic real-world data degradation, grouped into four categories: (1) Noise — Gaussian noise, shot noise, impulse noise; (2) Blur — defocus blur, glass blur, motion blur, zoom blur; (3) Weather — snow, frost, fog, brightness; and (4) Digital — contrast, elastic transformation, pixelation, JPEG compression. Each corruption type is offered at five severity levels, ranging from 1 (least severe) to 5 (most severe). In the main analysis, we employed severity level 1 to represent mild yet realistic distortions. Results for all severity levels are provided in the Supplementary Results. Robustness was measured using the mean corruption accuracy, calculated by averaging classification accuracy across all corruption types at a specific severity level. A higher mean corruption accuracy signifies that the model retains reliable performance across a broad range of corruptions, indicating improved generalization to distributional shifts.

\subsection{Adversarial robustness}

To assess the robustness of the neural network against adversarial attacks, we employed the Fast Gradient Sign Method (FGSM) \cite{goodfellow2014}. FGSM generates adversarial examples by adding perturbations in the direction that maximally increases the loss. Given an input $x$, label $y$, and loss function $\mathcal{L}(x, y)$, the adversarial example is constructed as:

\[
x_{\text{adv}} = x + \epsilon \cdot \text{sign}(\nabla_x \mathcal{L}(x, y)),
\]

where $\epsilon$ controls the magnitude of the perturbation. We evaluated adversarial robustness by varying $\epsilon$ from 0 to 0.05 and measuring the corresponding test accuracy on the perturbed inputs. A steep decline in accuracy with increasing $\epsilon$ indicates vulnerability to adversarial perturbations, whereas more stable accuracy suggests stronger robustness.

\section{Experimental details}

\subsection{Data availability}

The datasets used in this study are publicly available: \url{https://www.cs.toronto.edu/~kriz/cifar.html} (CIFAR-10 and CIFAR-100 \cite{krizhevsky2009}), \url{http://ufldl.stanford.edu/housenumbers/} (SVHN \cite{netzer2011}), \url{https://cs.stanford.edu/~acoates/stl10/} (STL-10 \cite{coates2011}).

\subsection{Code availability}

Python 3.12 (Python Software Foundation) with PyTorch 2.1 was used to perform the simulation. The code used in this work will be made available after the paper is published.

\subsection{Computing resources}

All simulations were performed on a computer with an Intel Core i9-13900K CPU and an NVIDIA GeForce RTX 4090 GPU. The simulation code was parallelized using PyTorch’s built-in parallelization to utilize the GPU resources efficiently.

\end{document}